\documentclass[journal]{IEEEtran}

\usepackage{times}
\usepackage{epsfig}
\usepackage{graphicx}
\usepackage{amsmath}
\usepackage{amssymb}

\usepackage{subfigure}

\usepackage{algorithm}
\usepackage{algorithmicx}
\usepackage{algpseudocode}
\usepackage{booktabs}
\usepackage{multirow}
\usepackage{color}
\usepackage{hyperref}

\usepackage{url}
\ifCLASSINFOpdf
\else
\fi
\begin{document}

%
\title{RelationRS: Relationship Representation Network for Object Detection in Aerial Images}
%
%
%

\author{~Zhiming~Liu,~Xuefei~Zhang,~Chongyang~Liu,~Hao~Wang,~Chao~Sun,~Bin~Li,~Weifeng~Sun,~Pu~Huang,~Qingjun~Li,~Yu~Liu,~Haipeng~Kuang,~Jihong~Xiu}

\maketitle

\begin{abstract}
Object detection is a basic and important task in the field of aerial image processing and has gained much attention in computer vision. However, previous aerial image object detection approaches have insufficient use of scene semantic information between different regions of large-scale aerial images. In addition, complex background and scale changes make it difficult to improve detection accuracy. To address these issues, we propose a relationship representation network for object detection in aerial images (RelationRS): 1) Firstly, multi-scale features are fused and enhanced by a dual relationship module (DRM) with conditional convolution. The dual relationship module learns the potential relationship between features of different scales and learns the relationship between different scenes from different patches in a same iteration. In addition, the dual relationship module dynamically generates parameters to guide the fusion of multi-scale features. 2) Secondly, The bridging visual representations module (BVR) is introduced into the field of aerial images to improve the object detection effect in images with complex backgrounds. Experiments with a publicly available object detection dataset for aerial images demonstrate that the proposed RelationRS achieves a state-of-the-art detection performance.
\end{abstract}


%
\IEEEpeerreviewmaketitle

\section{Introduction}
\label{sec:Introduction}
%
%
%
%
\IEEEPARstart{T}{he} technique of object detection in aerial images refers to extracting the positions of the objects defined by domain experts, according to whether a kind of object is common and its value for real-world applications \cite{DOTA,GLSNet}. With the rapid development of remote sensing acquisition technology, object detection methods in aerial images are widely used in ship monitoring, maritime search and rescue, traffic control, power line inspection, military reconnaissance, and other fields \cite{OcSaFPN}. In addition, for unmanned aerial vehicles (UAVs) and satellites with limited energy resources, aerial image object detection technology is also used as a pre-processing method to select important images with suspected targets and return them to the ground station first. This puts higher requirements on the accuracy and speed of object detection algorithms \cite{Lightweight-RS}. Based on this rapidly growing demand, traditional object detection algorithms using the logic of extracting hand-designed low-level features, and then classifying and predicting bounding boxes can no longer meet the detection tasks under fast, multiple categories and complex background conditions \cite{Optical Satellite Images With Complex Backgrounds,Ship Detection From Optical Satellite Images Based on Saliency Segmentation,A Survey}.

Widely used convolutional neural networks (CNNs) greatly promoted the development of object detection technology. Overall, the current object detection network can be divided into single-stage detectors and two-stage detectors. On the one hand, the two-stage detector method can be also called the region proposal-based method. This type of algorithm usually uses a small two-class network to extract the region proposals firstly, and then perform bounding boxes tuning procedure and category prediction. The region proposal-based method can achieve higher detection accuracy, but its structure is not suitable to edge computing devices. DPM \cite{DPM}, R-CNN \cite{RCNN}, Fast R-CNN \cite{FASTRCNN}, Faster R-CNN \cite{FASTERRCNN}, feature pyramid network (FPN) \cite{FPN}, Cascade R-CNN \cite{Cascade}, Mask R-CNN \cite{Mask r-cnn}, etc. are the representative algorithms. On the other hand, the single-stage detector can be also called the regression-based method, including Overfeat \cite{Overfeat}, you only look once (YOLO) \cite{YOLO}, YOLOv2 \cite{YOLOv2}, YOLOv3 \cite{YOLOv3}, YOLOv4 \cite{YOLOv4}, YOLOX \cite{YOLOX}, single shot detector(SSD) \cite{SSD}, FCOS \cite{Fcos}, and RetinaNet \cite{Focalloss}. The single-stage detector performs object detection on an entire image at one time, and predicts the bounding boxes as the coordinate regression task. This kind of method is very suitable for the embedding of edge computing devices. But its accuracy is not as good as the two-stage detector.

It is worth mentioning that some methods transform the prediction of the bounding boxes into a task of predicting the key points of the bounding boxes \cite{CornerNet,CornerNet-Lite,FoveaBox}. These methods are fast and efficient, but have difficulty in the scene where the object obscure each other in natural images.

\begin{figure}[t]
	\begin{center}
		\includegraphics[width=1.0\linewidth]{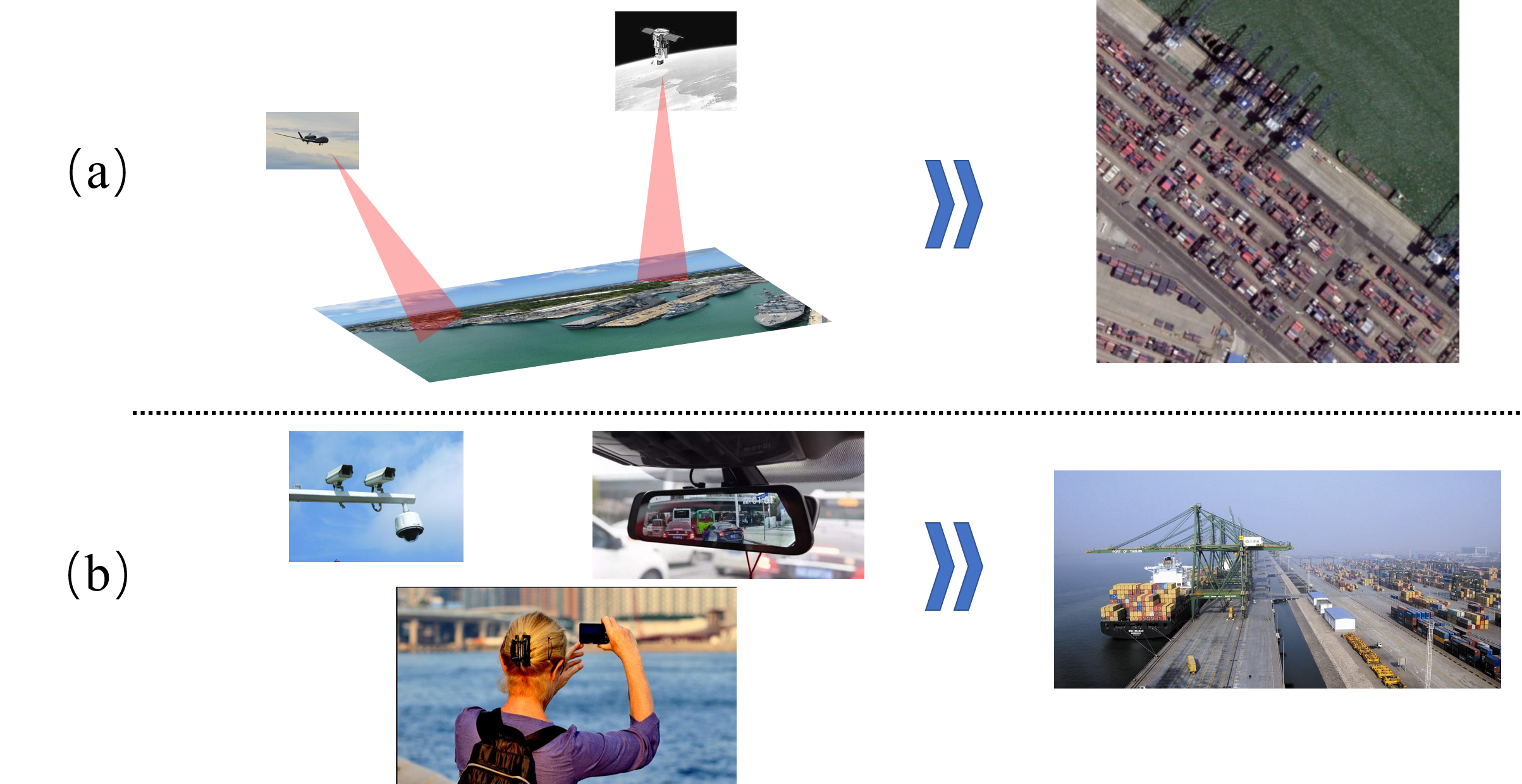}
	\end{center}
	\caption{The difference between natural images and aerial images. (\textbf{a}) A port image data obtained by aerospace platforms or aviation platforms. (\textbf{d}) A port image data captured by surveillance equipment or handheld devices.}
	\label{rs-and-natural}
\end{figure}

At present, encouraged by the great success of deep-learning-based object detection in natural images \cite{Deep learning,ResNet}, many researchers have proposed utilizing a similar methodology for object detection in aerial images \cite{A Survey,Ship Detection Visual Attention}. Figure~\ref{rs-and-natural} shows the difference between natural images and aerial images. First of all, the backgrounds of aerial images are more complex, which put forward higher requirements on algorithms. Secondly, the aerial images are taken from the vertical view perspective when acquiring, so there is almost no occlusion between the objects. In addition, aerial images have rich scene-target semantic information. Finally, due to the different acquisition platforms, flight trajectories, and sensors used when acquiring aerial images, almost every aerial image product has unique resolution and imaging characteristics. This leads to drastic scale changes of the same object. And the scales between different objects are also quite different. To sum up, it is impossible to directly apply the common object detection algorithms based on deep learning in the field of natural images to the field of aerial image object detection.

In this paper, in order to obtain improved object detection results with optical remote sensing images, we put forward an effective network structure, namely, relationship representation network for object detection in aerial images (RelationRS). In RelationRS, we first propose a multi-scale feature fusion module to deal with dramatic scale changes, which can dynamically learn the relationship of the object at different scales. Besides, the potential relationships between different scenes of aerial images can be learned by the multi-scale feature fusion module at the same time. The dual relationship module realizes the extraction and characterization of multi-scale relations and relations from different scenes, and dynamically generates weight parameters according to the input image data to guide the fusion of multi-scale features. According to the characteristic that the objects on the aerial image usually do not block each other, we then introduced the bridging visual representation module (BVR) \cite{BVR}, which integrates key-points detection and bounding box detection, into the field of aerial image object detection to meet the complex backgrounds. BVR can build relationships between different types of representations.

The proposed algorithm tested on the DOTA dataset \cite{DOTA} demonstrates that the proposed RelationRS achieves a state-of-the-art detection performance.

The rest of this paper is organized as follows. Section~\ref{sec:RelatedWork} gives a brief review of the related work on aerial image object detection based on deep learning, the multi-scale feature representations, and the conditional convolution mechanism. In Section~\ref{sec:ProposedMethod}, we introduce the proposed method in detail. The details of the dataset, the experiments conducted in this study, and the results of the experiments are presented in Section~\ref{sec:ExperimentsandResults}. Section~\ref{sec:Conclusions} concludes this paper with a discussion of the results.

\section{Related Work}
\label{sec:RelatedWork}
\subsection{Object Detection of Aerial Images}
\label{sec:ObjectDetectionofAerialImages}

Given the characteristics of aerial images, many researchers currently focus on the two-stage detector as the baseline to obtain higher-precision detection results, compared to the single-stage detector.

Zou et al. \cite{SVDNet} proposed a singular value decomposition network (SVDNet) for ship object detection in aerial images. To improve detection accuracy, the SVDNet combines singular value decomposition and convolutional neural networks. Since the region proposals are extracted by the selective search algorithm \cite{selective1,selective2}, this non-end-to-end implementation greatly reduces the speed of the algorithm. Dong et al. \cite{Sig-NMS} optimized the non-maximum suppression algorithm, and the proposed Sig-NMS method has better accuracy for small objects. Both Deng et al. \cite{Toward fast and accurate} and Xiao et al. \cite{Airport detection based on a multiscale} have used sliding windows to extract region proposals, then used convolutional neural networks to extract deep learning-based features, and finally used non-rotating rectangular boxes to characterize objects. A large number of regions in aerial images belong to the background class where no object exists. Therefore, the method of using the sliding windows to extract region proposals is inefficient, and it is easy to cause an imbalance between positive and negative samples. Xu et al. \cite{Deformable convnet with aspect ratio} and Ren et al. \cite{Small object based on Faster} applied deformable convolution \cite{Deformable} to aerial image object detection tasks with complex boundaries. Through expanding the receptive field of the convolution kernel, the deformable convolution with variable sampling position is beneficial to extract the complex boundaries of objects in aerial images. Similarly, Hamaguchi et al. \cite{Effective use of dilated convolutions} has used the dilated convolution filters \cite{dilated convolution} to optimize the receptive field size of the convolution kernel in aerial image object detection and segmentation, which is conducive to the extraction of scene semantic information. CDD-Net \cite{CDD-Net} has used an attention mechanism to improve the ability of multi-scale feature representation. In addition, CDD-Net has used a local semantic feature network to extract the feature information of the area around an object, to compensate for the limited filter size of the CNNs. Finally, the detection task is finished with both the feature maps of region proposals and the information around objects. To improve the efficiency of multi-scale feature utilization, A2RMNet \cite{A2RMNet} adopted a gate structure to automatically select and fuse multi-scale features and then adopted the same region proposal feature map pooling method as R2CNN \cite{R2CNN} to obtain three different aspect ratios. Feature maps and the attention mechanism are also used to optimize and merge the three feature maps to improve the accuracy of object detection.

Different from the algorithms using the rectangular bounding box frame, some aerial image object detection methods based on two-stage detectors believe that the use of the rotating bounding box frame can better describe the object and reduce the influence of background pixels. Although RRPN \cite{RRPN} is proposed for the task of text detection, it has achieved excellent results in many competitions related to aerial imagery. Similarly, both Li et al. \cite{Rotation-insensitive} and Ding et al. \cite{RoITransformer} adopted the lightweight structure to obtain more accurate rectangular bounding box characterization with different angles in the region proposal extraction stage. When performing RoI Pooling operations, this kind of rotation bounding box frame tries to eliminate the background pixels in the region proposal to improve the characteristics of the object itself. Yang et al. \cite{Scrdet} constructed a two-stage detector, which performs well in small objects and densely arranged objects detection tasks, namely SCRDet. Firstly, for SCRDet, a new feature fusion branch, called SF-Net, is proposed to improve the recall value of the region proposals. Then, SCRDet adopted a MDA-Net with self-attention mechanism to reduce the interference of background pixels and improve the feature representation ability of the object itself. Finally, the loss function is improved with a constant factor. Both CAD-Net \cite{CAD-Net} and GLS-Net \cite{GLSNet} improved feature representation ability of the object itself by global semantic branches with attention mechanism or saliency algorithm. These global semantic branches are used to compensate for the lack of scene information caused by the limited size of the receptive field.

RADet \cite{RADet} improved the feature representation ability for scale changes by fusing the front layer feature maps and the deeper layer feature maps, this kind of fusion strategy is useful for the small object detection tasks. Moreover, on the basis of on Mask R-CNN \cite{Mask r-cnn}, RADet can predict the instance mask and the rotating rectangular bounding box at the same time based on the attention mechanism. In RADet, the minimum rectangular area of the object is used as the truth of the mask, which actually still contains the background area. For this reason, RADet does not realize the high-precision characterization of the object itself. Similarly, inspired by Mask R-CNN, Mask OBB \cite{Mask OBB} adopted the inception module \cite{Inception-v4} to fuse feature maps from different depths and utilized the attention mechanism to construct semantic segmentation feature maps. Finally, Mask OBB can predict the bounding boxes, rotated bounding boxes, and instance masks simultaneously. In order to deal with the problem of scale changes and small objects, Li et al. \cite{Learning object-wise semantic} proposed a method that can predict bounding boxes and rotated bounding boxes with a module similar to the inception module and a semantic segmentation module. However, the above two methods require multiple types of samples in training stage. Xu et al. \cite{Gliding vertex} believe that it is difficult to directly predict the rotated bounding boxes, and the detection of the rotated objects should to be implemented step by step. That is, the non-rotated bounding boxes need to be predicted first, and then the offset of the four vertices can be calculated. Zhu et al. \cite{Adaptive period embedding} proposed that it is difficult and inefficient to directly calculate the angles of the bounding boxes. Therefore, two different two-dimensional periodic vectors are used to represent angles, and a new intersection over union (IoU) is used to solve the problem of the object with a large length and width ratio. Fu et al. \cite{Rotation-aware and multi-scale} extracted features of region proposals with angles and merged feature maps from different depths through bidirectional information flow to enhance the representation ability of the feature pyramid network. ReDet \cite{Redet} encodes the rotation-equivariance and the rotation-invariance, which can reduce the demand for network parameters while realizing the detection of rotated objects. Based on the characteristics of images in frequency, OcSaFPN \cite{OcSaFPN} is proposed to improve the accuracy of object detection in aerial images with noise.

Object detection algorithms based on single-stage detectors in aerial images have developed rapidly, and the accuracy gap with two-stage detectors has been narrowing or even surpassing. The aerial image object detection algorithms based on the single-stage detectors have the common characteristics of being fast, concise, and conducive to the deployment of dedicated computing chips and edge computing equipment. For this reason, object detection algorithms based on single-stage detectors in aerial images have great potential in the production and application in a variety of scenarios. By adding a new branch for scene prediction, you Only Look Twice (YOLT) \cite{YOLT} realizes the simultaneous prediction of the objects and scene information, and forms the association between the object and the scene information from the loss function. But this method does not make up for the shortcomings of YOLOv2 itself, and most of the datasets in aerial image field lack scene labeling information. Inspired by SSD, FMSSD \cite{FMSSD} adopted the dilated convolutional filter to enlarge the size of the receptive field. Although this method can extract the information of the object and its surrounding area at the same time, the size of the surrounding area is limited and the feature maps still cannot describe the scene semantics well. Zou et al. \cite{Random access memories} realized the small objects detection in high-resolution remote sensing images based on Bayesian priors algorithm, and optimized the memory overhead when processing large images. On the basis of RetinaNet, R3Det \cite{R3det} predicts the rotated bounding boxes through the anchor with angles and the first head network. In addition, a feature refinement module (FRM) is designed to reconstruct the entire feature map to solve the problem of misalignment. The FRM has a lightweight structure, rigorous and efficient code implementation, and can be easily inserted into a variety of cascaded detectors.

\subsection{Multi-Scale Feature Representations}
\label{sec:Multi-ScaleFeatureRepresentations}

For convolutional neural networks, the fusion and use of multi-scale feature maps can greatly improve the detection accuracy of the algorithm in small target detection tasks and scenes with dramatic scale changes \cite{OcSaFPN}.

In the field of object detection based on convolutional neural networks, feature pyramid network (FPN) \cite{FPN} is one of the earliest effective ways to solve multi-scale problems, and it is also the most widely used feature pyramid structure. FPN receives the multi-scale features from the backbone structure and then builds a top-down information transfer path to enhance the representation capabilities of the multi-scale features. While retaining the top-down information flow path of FPN, PANet \cite{PANet} adds a bottom-up information transmission path to realize the two-way interaction between shallow features and deep features. Furthermore, the adaptively spatial feature fusion (ASFF) \cite{ASFF} is designed with dense connections to transfer information between features of different scales.

The scale-transfer module proposed by STDL \cite{STDL} reconstructs multi-scale feature maps without introducing new parameters. Kong et al. \cite{Deep feature pyramid reconfiguration} first fused multi-scale feature maps and then used a global attention branch to reconstruct these features. Both AugFPN \cite{Augfpn} and $U^{2}-ONet$ \cite{U2-ONet} output multiple feature maps of different scales, and then perform loss calculations on each level.

NAS-FPN \cite{Nas-fpn} adopted an adaptive search algorithm to allow the system to automatically find the optimal multi-scale feature information flow path, thereby forming a pyramid structure with a fixed connection path. This kind of network design logic is different from the common feature pyramid, which can avoid the complicated manual design process, but the automatically search process requires huge computing resources. Inspired by NAS-FPN, MnasFPN \cite{Mnasfpn} added the characteristics of mobile hardware to the search algorithm. Therefore, when searching for the optimal network structure, the search algorithm not only considers accuracy as the only basis for judgment but also takes the hardware characteristics into account. Therefore, the deployment of MnasFPN on mobile is more advantageous.

BiFPN \cite{BiFPN} improved multi-scale expression ability through connections across different scales and short-cut operations. OcSaFPN \cite{OcSaFPN} improved the robustness of multi-scale features on noisy data by assigning different weights to feature maps from different depths.

\subsection{Conditional Convolution Mechanism}
\label{sec:ConditionalConvolutionMechanism}

Conditional convolution, which can also be called dynamic filter, was first proposed by Jia et al. \cite{Dynamic filter}. Different from the traditional convolutional layers with fixed weight parameters in the inference stage, the parameters of a conditional convolutional layer are constantly changing with different input data. Therefore, the parameter form is more flexible. This variable parameter form can better adapt to the input data, thus it has gradually attracted the attention of many researchers.

At the same time, hyper networks \cite{Hypernetworks} is proposed to generate weights for another network. Later, this mechanism is also adopted to the style transfer task \cite{Neural style transfer}. A kind of dynamic upsampling filter is used by Jo et al. \cite{Deep video super-resolution} for the task of high-resolution video reconstruction. Similarly, Meta-SR \cite{Meta-SR} is also adopted to the super-resolution reconstruction task. Condconv \cite{Condconv} described the logic of conditional convolution in detail and used the form of group convolution to deal with the situation of multiple data in a batch. Wu et al. \cite{Dynamic filtering with large sampling} the idea of the dynamic filter to generate optical flow data. Both Harley et al. \cite{Segmentation-aware convolutional networks} and CondInst \cite{CondInst} adopted the conditional convolution mechanism to predict the instance masks. Xue et al. \cite{Visual dynamics} adopted the conditional convolution mechanism to generate the future frames. Both Sagong et al. \cite{cGANs} and Liu et al. \cite{Learning to predict layout-to-image} introduced the conditional convolution mechanism into the generative adversarial network. CondLaneNet \cite{CondLaneNet} and ConDinet++ \cite{ConDinet++} are respectively for the lane line extraction task and the road extraction task in remote sensing images.

In summary, most of the current researches on conditional convolution focus on pixel-level tasks, and the number of related researches is generally small. Applications related to remote sensing images are even rarer.

\section{Proposed Method}
\label{sec:ProposedMethod}

\begin{figure*}
	\begin{center}
		\includegraphics[width=0.8\linewidth]{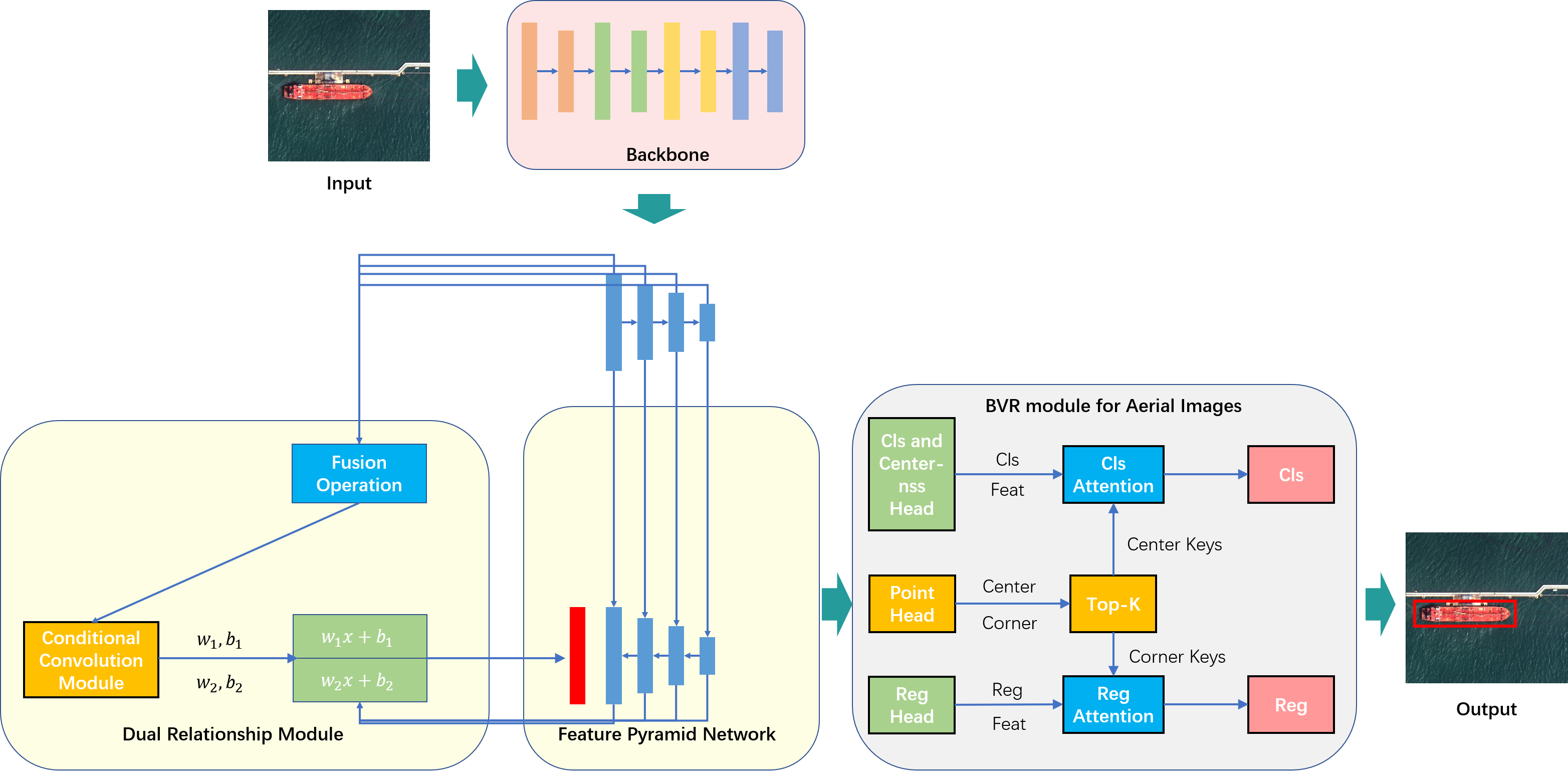}
	\end{center}
	\caption{The proposed framework, namely, the relationship representation network for object detection in aerial images (RelationRS), is made up of three components: the baseline network, which is made up of the fully convolutional single-stage detector with no anchor setting (FCOS) \cite{Fcos} and the feature pyramid network (FPN) \cite{FPN}; the dual relationship module, which learns the potential relationship between different scales of the objects and the potential relationship between different scenes of aerial images in one batch; the bridging visual representation module for aerial image object detection task, which learns the potential relationship between different coordinate representations based on BVR module \cite{BVR}.}
	\label{RelationRS}
\end{figure*}

\begin{figure}[t]
	\begin{center}
		\includegraphics[width=1.0\linewidth]{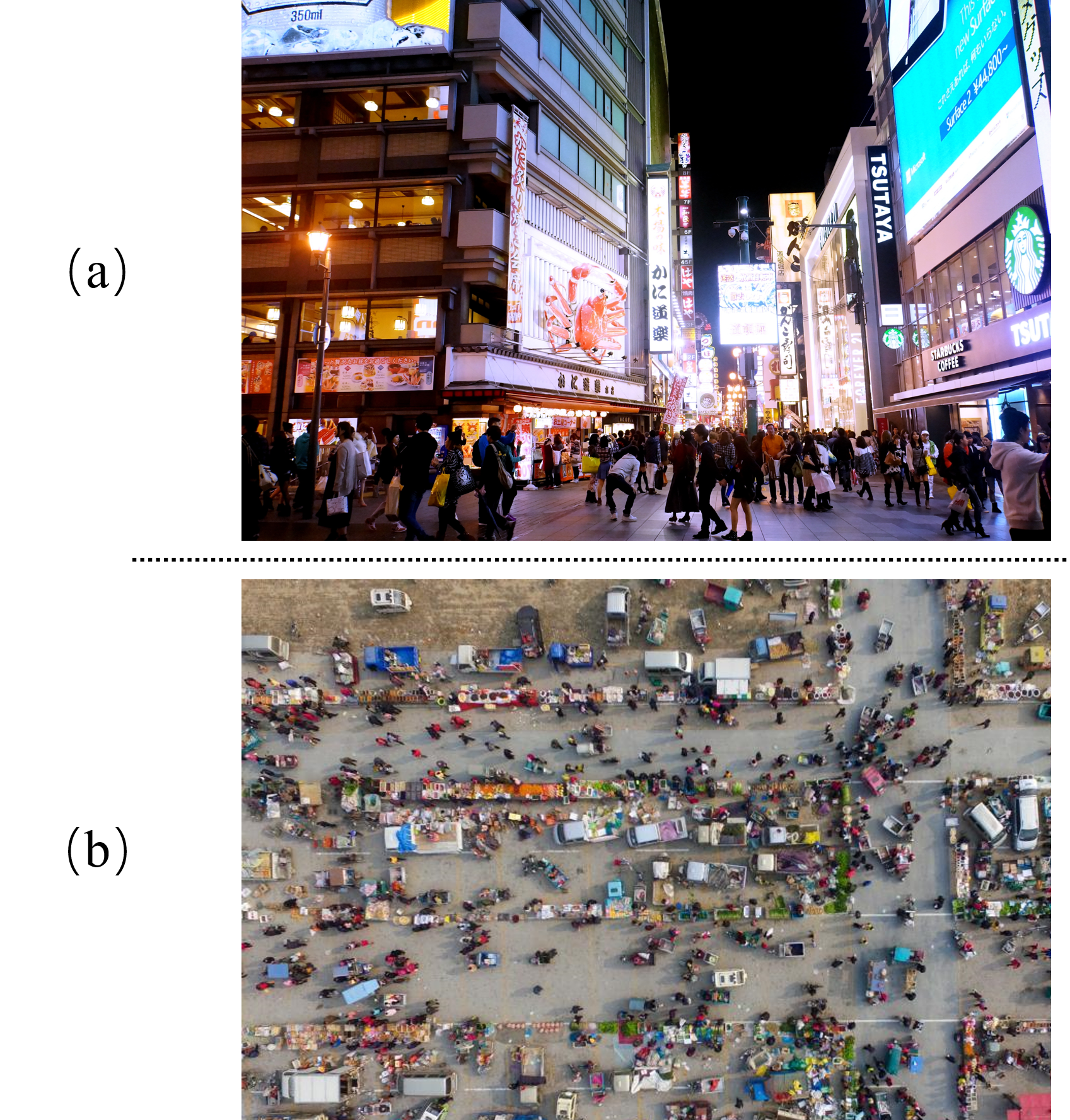}
	\end{center}
	\caption{Comparison of target occlusion in images under different viewing angles. (a)There are serious occlusion problems between different people in natural images. (b)Images acquired by drones, due to the overhead perspective, there is almost no problem of mutual occlusion between people and vehicle objects.}
	\label{OcclusionProblem}
\end{figure}

In this section, we introduce the RelationRS algorithm. The flow chart of the proposed object detection method is shown in Figure~\ref{RelationRS}. The proposed RealationRS is based on the classic anchor-free detector, namely FCOS \cite{Fcos}, with the backbone module (ResNet50 \cite{ResNet}) and FPN structure \cite{FPN}. Firstly, the input aerial image data flows through the backbone network and FPN module, and then feature maps of five scales can be obtained (${P_{2}, P_{3}, P_{4}, P_{5}, P_{6}}$). In order to better adapt to multi-scale object detection tasks, the dual relationship module is designed to fuse features of different scales, which is explained in Section~\ref{sec:DualRelationshipModule}. On the one hand, this dual relationship module can learn the relationship of an object at different scales, and dynamically generate the fusion weights according to the input data to guide the fusion of multi-scale information. On the other hand, the dual relationship module can learn the potential scene semantics between different patches in one batch, and improve the detection accuracy through the comparison between different scenes. In Section~\ref{sec:BridgingVisualRepresentationsforObjectDetectioninAerialImages}, on the basis of fusion of multi-scale information, we use the bridging visual representations module to suppress the influence of complex background information in aerial images, and improve accuracy through the combination of multiple features. Aerial imagery usually adopts top-view perspective imaging, thus there is little occlusion between objects (Figure~\ref{OcclusionProblem}). One of the disadvantages of the key point-based object detection algorithm is that it is not robust enough when encountering occlusion problems, and its advantage lies in better positioning accuracy. Based on the above reasons, the use of key-point detection technology in aerial imagery can achieve strengths and avoid weaknesses. By combining rectangular bounding box detection, center detection, corner detection, and classification, the interference of complex background information can be suppressed, and the positioning accuracy of objects on complex background data can be improved. Finally, high-precision aerial image object detection can be achieved.

\subsection{Dual Relationship Module (DRM)}
\label{sec:DualRelationshipModule}

\begin{figure}[t]
	\begin{center}
		\includegraphics[width=1.0\linewidth]{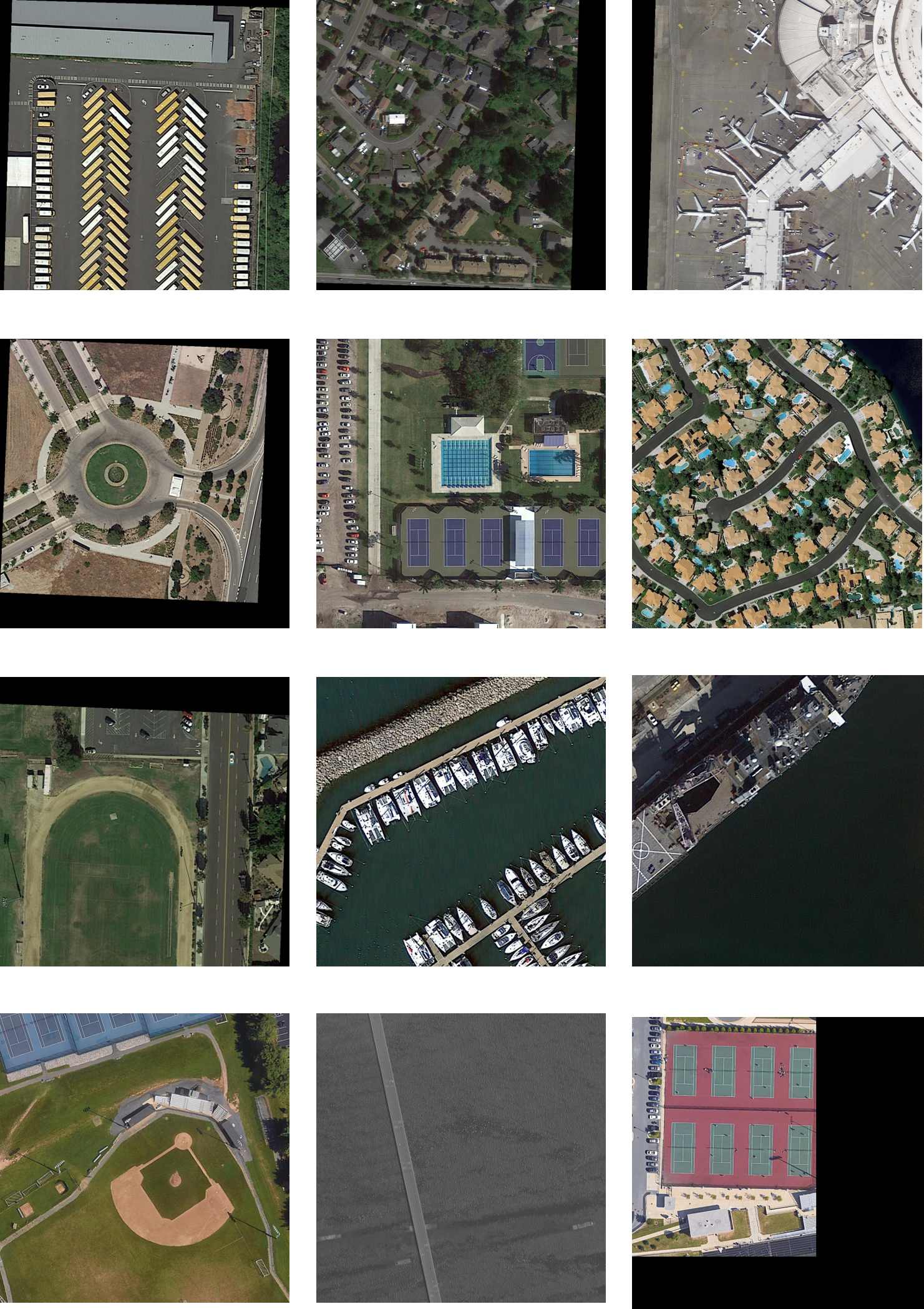}
	\end{center}
	\caption{Sample aerial image patches with different scenes. The sizes of patches are all $1024 \times 1024$. Different patches have different scene semantics, forming a potential semantic contrast with each other. There are also intra-class and inter-class scale differences for objects between different scenarios.}
	\label{MultipleScenes}
\end{figure}

Aerial images have obvious characteristics of diverse scales and the existence of scene semantics. To deal with the scale changes within and between classes, FPN \cite{FPN}, PANet \cite{PANet}, NAS-FPN \cite{Nas-fpn}, MnasFPN \cite{Mnasfpn}, BiFPN \cite{BiFPN}, OcSaFPN \cite{OcSaFPN}, etc. have all been proposed. These methods effectively improve the multi-scale object detection problem. However, the structures and weights of these methods are fixed in the inference stage and will not change according to the input data. As shown in Figure~\ref{MultipleScenes}, for different aerial image patches, the semantic information of the scene contained in it is different, and the object types and scales in the two scenes are also quite different. Based on the above reasons and inspired by CondInst \cite{CondInst}, the dual relationship module is designed to learn scale changes from multi-scale information, implicitly extract the connections and differences between the scenes contained in different patches in the one batch. In addition, the neural network parameters of the multi-scale information fusion module are dynamically generated to guide the fusion of multi-scale features with semantic information of the input data.

Figure~\ref{DualRelationshipModule} shows the construction of the dual relationship module. Taking $batchsize=2$ as an example, the four scale feature maps extracted from the backbone network are marked as ${C_{2}, C_{3}, C_{4}, C_{5}}$. Then after taking ${C_{2}, C_{3}, C_{4}, C_{5}}$ as input, the feature maps output by FPN \cite{FPN} can be marked as ${P_{2}, P_{3}, P_{4}, P_{5}, P_{6}}$. Among them, $P_{6}$ is generated by $P_{5}$ with the maximum pooling operation. Inspired by CondInst, the key point of the dual relationship module is the generation of $P_{2}^{\prime}$ feature maps, and the generation process of $P_{2}^{\prime}$ can be described by Equation~(\ref{eq:DualRelationshipModule}):

\begin{figure}[t]
	\begin{center}
		\includegraphics[width=1.0\linewidth]{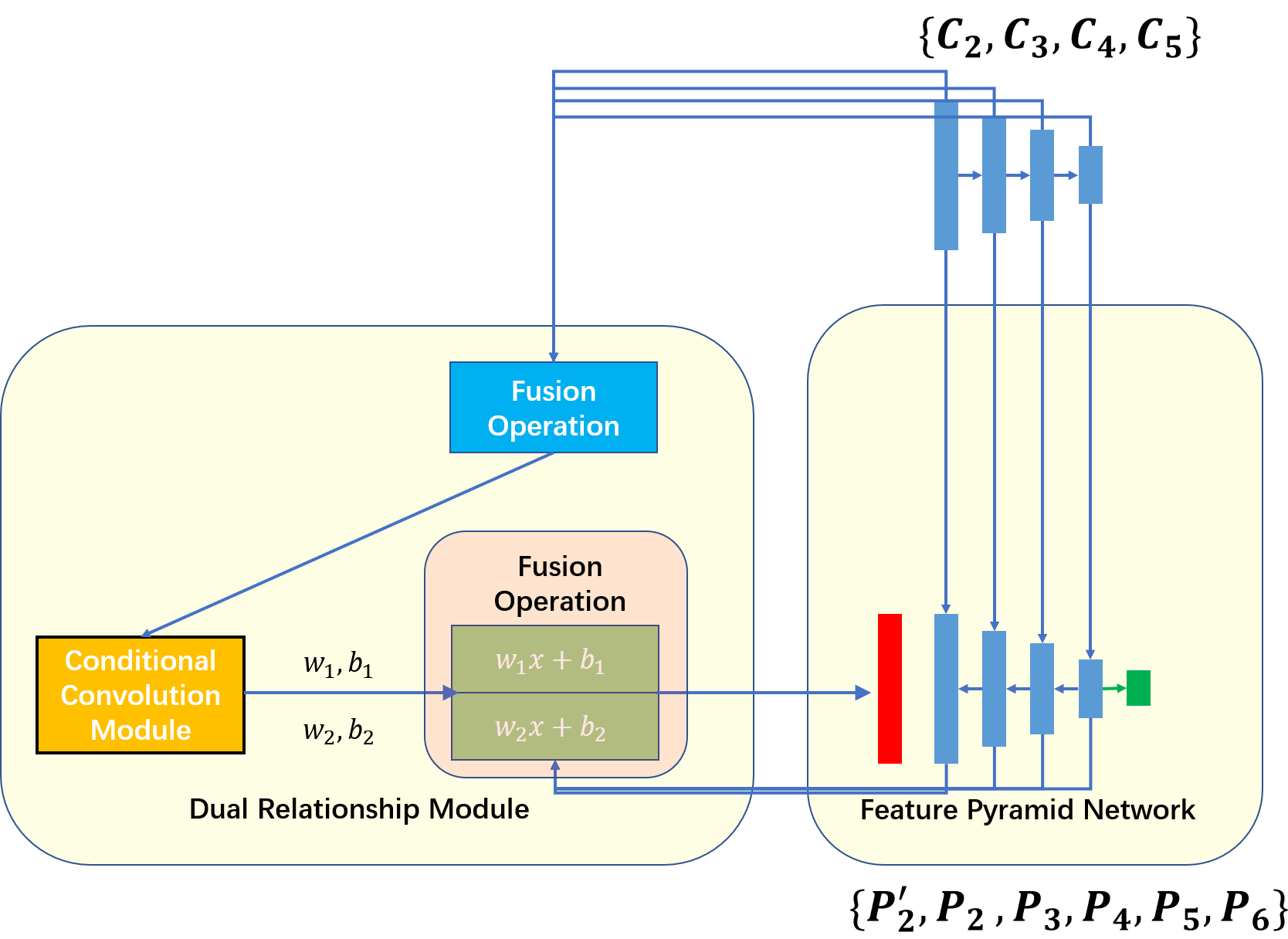}
	\end{center}
	\caption{The construction process for the dual relationship module.}
	\label{DualRelationshipModule}
\end{figure}

\begin{equation}
P_{2}^{\prime} = Conv2(Conv1(concate(resize((P_{2}, P_{3}, P_{4}, P_{5}))))),
\label{eq:DualRelationshipModule}
\end{equation}
where $Conv1(\cdot)$ and $Conv2(\cdot)$ denote two convolution operation with kernel size [1, 1], $concate(\cdot)$ denotes the concatenation. The parameters of $Conv1(\cdot)$ and $Conv2(\\cdot)$ are obtained by CondConv \cite{Condconv}, and the process can be expressed by Equation~(\ref{eq:CondConv}):

\begin{equation}
w_{1},w_{2},b_{1},b{2} = CondConv(FO(C_{2}, C_{3}, C_{4}, C_{5})),
\label{eq:CondConv}
\end{equation}
where $CondConv(\cdot)$ denotes the conditional convolution module from CondConv \cite{Condconv}, $FO(\cdot)$ denotes the fusion operation seen in Figure~\ref{DualRelationshipModule} and can be built by Equation~(\ref{eq:FusionOperation}):

\begin{equation}
FO(C_{2}, C_{3}, C_{4}, C_{5}) = Conv_{3 \times 3}(concate(resize(C_{2}, C_{3}, C_{4}, C_{5}))),
\label{eq:FusionOperation}
\end{equation}
where $Conv_{3 \times 3}(\cdot)$ denotes the convolution operation with kernel size [3, 3].

Based on the above formulas, ${C_{2}, C_{3}, C_{4}, C_{5}}$ are resized and concated in the channel dimension. In order to obtain $w_{1},w_{2},b_{1},b{2}$, the feature map after fusion first passes through a convolutional layer with kernel size [3, 3] for feature alignment and channel dimensionality reduction. Then, the dimensionality-reduced feature map is sent to CondConv module to generate the required weights and bias values. To solve the situation where batchsize is not equal to 1, we treat different patches in a batch as different experts. This is different from the way in CondInst. Finally, we generate the parameters to initialize the two convolutional layers for the fusion of ${P_{2}, P_{3}, P_{4}, P_{5}}$, and finally output the required feature maps $P_{2}^{\prime}$. In this way, we get a series of multi-scale feature maps ${P_{2}^{\prime}, P_{2}, P_{3}, P_{4}, P_{5}, P_{6}}$ and send them to the head network. To reduce the amount of parameters ($w_{1},w_{2},b_{1},b{2}$), the group convolution mechanism from AlexNet \cite{AlexNet} is used in $Conv1(\cdot)$ and $Conv2(\cdot)$.

\begin{figure}[t]
	\begin{center}
		\includegraphics[width=1.0\linewidth]{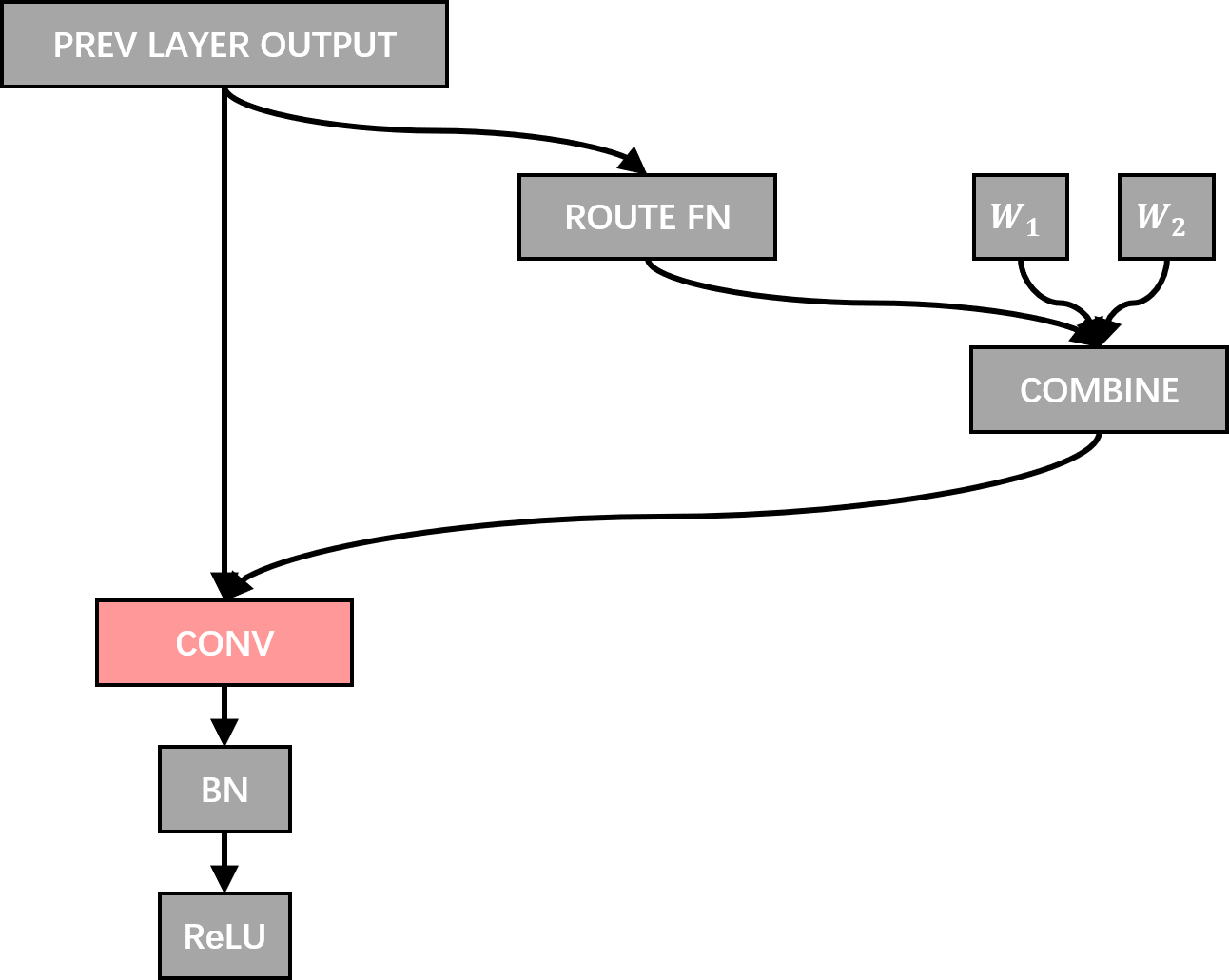}
	\end{center}
	\caption{CondConv \cite{Condconv} layer architecture with $n=2$ kernels. CondConv: $({\alpha}_{1}W_{1}+{\alpha}_{2}W_{2})\times x$.}
	\label{CondConv}
\end{figure}

It is worth noting that we treat two patches in a batch as experts (Figure~\ref{CondConv}), and then generate weight parameters. These parameters potentially obtain the relationships and differences between two scenarios, thus the network can dynamically extract features based on the input data. This is different from the idea of using semantic extraction branches in CAD-Net \cite{CAD-Net} and GLS-Net \cite{GLSNet} to extract the semantics of a single patch scene.

\subsection{Bridging Visual Representations for Object Detection in Aerial Images}
\label{sec:BridgingVisualRepresentationsforObjectDetectioninAerialImages}

Generally speaking, different representation methods usually lead detectors to perform well in different aspects. According to the structures of the current object detection networks, the two-stage detector can usually obtain a more accurate object category prediction. The detection method based on the center point can improve the detection accuracy of small objects. The corner-based method reduces the characterization dimension of the bounding boxes, thus it has more advantages in positioning tasks \cite{BVR}. As shown in Figure~\ref{OcclusionProblem}, in the face of the character that objects in aerial images are less occluded, we believe that the introduction of the key-point detection algorithm based on FCOS can suppress the influence of complex backgrounds and improve detection accuracy. Based on this point of view, the BVR module is introduced into the aerial image object detection task. Through the combination of multiple characterization methods, the accuracy of aerial image detection tasks for complex backgrounds can be improved.

For a detector, the main idea of BVR module is to regard its main representation as the master representation. Thus other auxiliary representations are adopts to enhance the master representation by a kind of transformer module \cite{Transformer}. That is, the transformer mechanism is used to bridge different visual representations.

\begin{figure}[t]
	\begin{center}
		\includegraphics[width=1.0\linewidth]{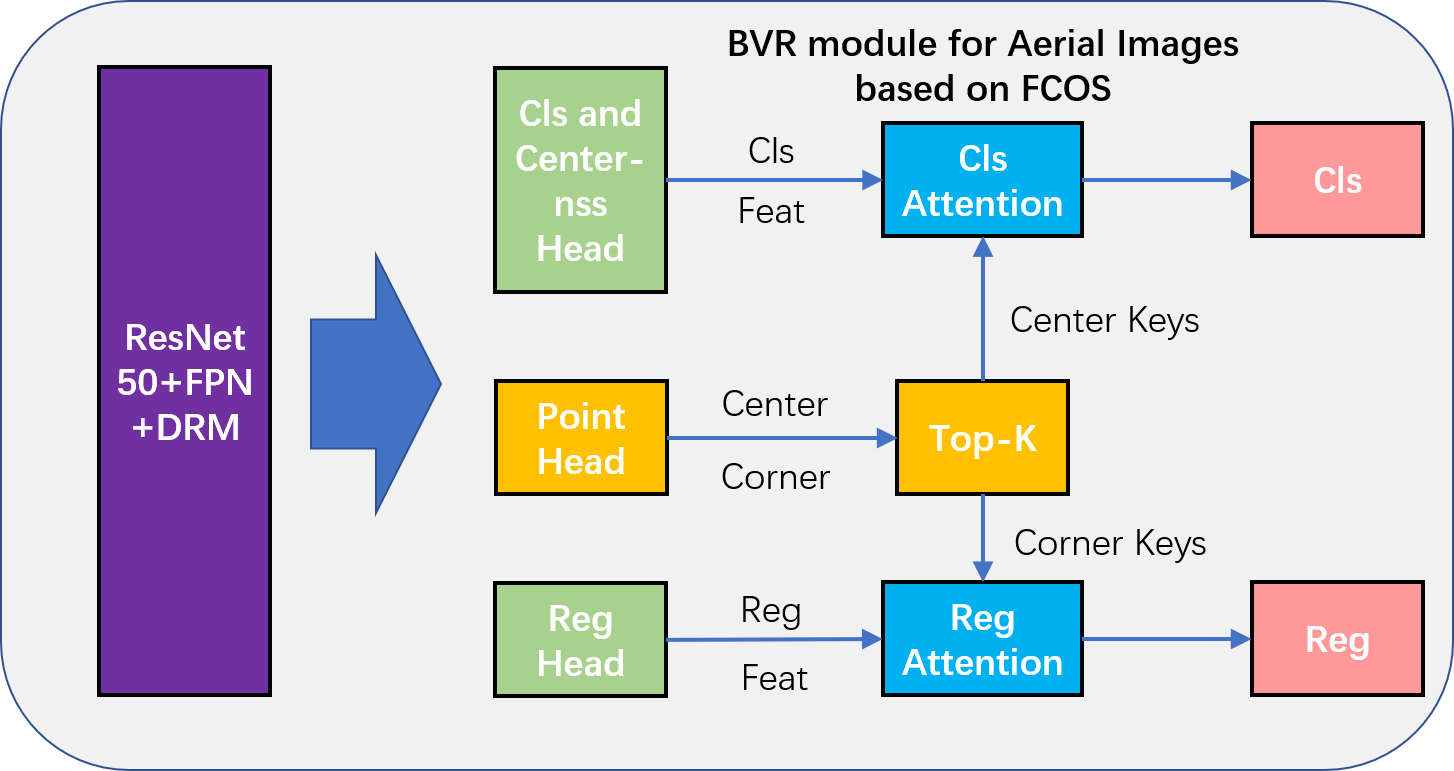}
	\end{center}
	\caption{The construction process for the BVR module based on FCOS.}
	\label{BVR_FCOS}
\end{figure}

As for the anchor-free algorithm (FOCS), the center point location and the corresponding features are regarded as the master representation and the query input at the same time. Compared with standard FOCS head network, the BVR module constructs an additional point head network (Figure~\ref{BVR_FCOS}). The point head network consists of two shared convolutional layers with kernel size [3, 3], followed by two independent sub-networks to predict the scores and sub-pixel offsets for center and corner prediction \cite{BVR}. The representations produced by the point head network are regarded as the auxiliary representation and the keys in the transformer algorithm. To reduce the amount of calculation, a top-k key selection strategy is adopted to control the set of keys not larger than $k$ (default=50), according to their corner-ness scores. Besides, the cosine/sine location embedding algorithm is used to reduce the complexity of coordinate representations. Here, according to the characteristic of the aerial images, the maximum number of key points is set to 400.

Based on the above settings of the queries and the keys, the enhanced features $f_{i}^{\prime q}$ can be calcuted by Equation~(\ref{eq:attention}):

\begin{equation}
f_{i}^{\prime q} = f_{i}^{q} + \sum_{j}S(f_{i}^{q},f_{j}^{k},g_{i}^{q},g_{j}^{k})\cdot T_{v}(f_{j}^{k}),
\label{eq:attention}
\end{equation}
where $f_{i}^{q}$ and $g_{i}^{q}$ are the input feature and geometric vector for a $query$ instance $i$; $f_{j}^{k}$ and $g_{j}^{k}$ are the input feature and geometric vector for a $key$ instance j; $T_{v}(\cdot)$ is a linear $value$ transformation function; $S(\cdot)$ is a similarity function between $i$ and $j$ \cite{BVR}. $S(\cdot)$ can be described as Equation~(\ref{eq:similarityfunction}):

\begin{equation}
S(f_{i}^{q},f_{j}^{k},g_{i}^{q},g_{j}^{k}) = {softmax}_{j}(S^{A}(f_{i}^{q},f_{j}^{k})+S^{G}(g_{i}^{q},g_{j}^{k})),
\label{eq:similarityfunction}
\end{equation}
where $S^{A}(f_{i}^{q},f_{j}^{k})$ denotes the appearance similarity computed by a scaled dot product between $query$ and $key$ features,and $S^{G}(g_{i}^{q},g_{j}^{k})$ denotes a geometric term computed by cosine/sine location embedding-based method \cite{BVR}.

The BVR module based on FCOS can be seen in Figure~\ref{BVR_FCOS}.

\section{Experiments and Results}
\label{sec:ExperimentsandResults}

For proving the effectiveness of the proposed method, a widely used ``A Large-Scale Dataset for Object Detection in Aerial images” (DOTA) \cite{DOTA} dataset was used in the experiments for the object detection task in aerial images. In this chapter, the DOTA1.0 dataset, the implementation details, and the ablation studies conducted with the proposed method can be introduced in detail in order.

\subsection{Dataset}
\label{sec:Dataset}

\textbf{DOTA1.0}. DOTA1.0 dataset is one of the largest published open-access dataset for object detection in aerial images. The dataset consists of 2806 large-size aerial images from Google Earth and satellites including Julang-1 (LJ-1) and the Gaofen-2 satellite (GF-2). The dataset consistes 188,282 annotated bounding boxes in 15 categories, including plane, baseball diamond (BD), bridge, ground~ track field (GTF), small vehicle (SV), large vehicle (LV), ship, tennis court (TC), basketball court (BC), storage tank (ST), soccer-ball field (SBF), roundabout (RA), harbor, swimming pool (SP), and~ helicopter (HC). Most of the ground sampling distance(GSD) values of the images in the DOTA1.0 dataset are better than 1 meter. The DOTA1.0 dataset has the characteristics of diverse scenarios, categories, scales, etc. It is still challenging to achieve high-precision object detection with this dataset.

In the experiments, the training set and validation set of DOTA1.0 were used in the training stage, and the test set was used for the inference stage and evaluation stage. The original images were all cropped to a size of $1024 \times 1024$, overlapping by 500 pixels. If the size of a patch is less than $1024 \times 1024$, the zero padding method was adopted for completion. Based on the above settings, we obtained a total of 38,504 patches for training stage and a total of 20,012 patches for the evaluation task. Since the ground truth files of the test set from the DOTA1.0 dataset was not disclosed, we submitted the final test results to the online evaluation website in the format of '.txt' (\url{https://captain-whu.github.io/DOTA/evaluation.html}).

\subsection{Evaluation Metrics}
\label{sec:EvaluationMetrics}

For quantitative accuracy evaluation, the mean average precision (mAP) is used in this paper. The mAP describes the mean value of the average precision (AP) values for multiple categories in a dataset. For a certain category, the AP value is the area enclosed by the coordinate axis and the broken line in the corresponding precision-recall graph. The larger the area, the higher its AP value. The details of the evaluation follow the official DOTA1.0 evaluation website (\url{https://captain-whu.github.io/DOTA/evaluation.html}).

\subsection{Implementation Details}
\label{sec:ImplementationDetails}

For the realization of the RelationRS, we built the baseline network based on FCOS \cite{Fcos} with FPN \cite{FPN}. The ResNet50 \cite{ResNet} pretrained on ImageNet \cite{ImageNet Classification} was adopted as the backbone network. A series of experiments were designed to better evaluate the effects of the dual relationship module and the bridging visual representations module for  aerial image object detection in this paper. The environment used was a single NVIDIA Tesla V100 GPU with 16 GB memory, along with the PyTorch 1.8.0 and Python 3.7.10 deep learning frameworks. The initial learning rate was 0.0025, the batch size of the input data was $2$, the value of the momentum is 0.9, the value of the weight decay was 0.0001, and the minibatch stochastic gradient descent (SGD) was also used for optimization. And the project was build on the mmdetection v2.7.0 \cite{MMDetection}.

\subsection{Ablation Experiments}
\label{sec:AblationExperiments}

Two ablation experiments were used to further discuss the influence of the dual relationship module and the bridging visual representations module. Here, the abbreviations in the DOTA data set are explained again: plane, baseball diamond (BD), bridge, ground track field (GTF), small vehicle (SV), large vehicle (LV), ship, tennis court (TC), basketball court (BC), storage tank (ST), soccer-ball field (SBF), roundabout (RA), harbor, swimming pool (SP), and helicopter (HC).

\subsubsection{Dual Relationship Module}
\label{sec:AblationDualRelationshipModule}

We conducted the ablation experiment to verify the effectiveness of the proposed dual relationship module. The baseline is the FCOS algorithm. +DRM means the combination of the FCOS and the dual relationship module. The difference between the baseline and the +DRM is only whether the dual relationship module is additionally used. And the parameters used in the experiment are strictly kept consistent.

From Table~\ref{tb:plusDRM}, +DRM obtains a $mAP$ of $65.63\%$, which is $1.38\%$ higher than the $mAP$ value of the baseline ($64.25\%$). For DOTA datasets with 15 categories, the baseline method only has advantages in three categories, plane, soccer-ball field (SBF) and basketball court (BC), which are $0.17\%$, $4.21\%$ and $5.4\%$ higher than the values of +DRM. This indicates that the performance of +DRM is not stable enough for objects with relatively large scales. For small objects, +DRM has achieved better accuracy in multiple categories. The $AP$ values of small vehicle (SV), large vehicle (LV), ship, storage tank (ST), and helicopter (HC) of the +DRM outperform the values of the baseline by $1.66\%$, $3.07\%$, $4.97\%$, $6.48\%$, and $6.46\%$. In addition, The $AP$ values of baseball diamond (BD), bridge, ground track field (GTF), roundabout (RA), harbor and swimming pool (SP) of the +DRM are also higher than values of the baseline. Therefore, the +DRM method can effectively improve the detection accuracy of small targets.

\begin{table*}[]
\caption{Detection accuracy in the ablation study of using DRM or not with DOTA test dataset. The bold numbers denote the highest values in each class.}
\centering
\resizebox{\textwidth}{!}{
\begin{tabular}{cccccccccccccccccccc}
\toprule
 \multirow{2}{*}{\textbf{Method}} & \textbf{Plane}& \textbf{BD}& \textbf{Bridge}& \textbf{GTF}& \textbf{SV}& \textbf{LV}& \textbf{Ship}& \textbf{TC} & \multirow{2}{*}{\textbf{mAP(\%)}}\\
 \cline{2-9}
 ~ & \textbf{BC} & \textbf{ST}& \textbf{SBF}& \textbf{RA}& \textbf{Harbor}& \textbf{SP}& \textbf{HC}\\
 \midrule
 \multirow{2}*{baseline} & \textbf{88.12}& 70.77& 44.04& 47.46& 76.36& 65.34& 77.96& 90.83&\multirow{2}{*}{64.25}\\ 					
 \cline{2-9}
 ~ & \textbf{74.31}& 78.37& \textbf{48.3}& 52.62& 72.25& 42.77& 34.27\\ \hline
 \multirow{2}*{+DRM} & 87.95& \textbf{71.66}& \textbf{44.1}& \textbf{52.48}& \textbf{78.02}& \textbf{68.41}& \textbf{82.93}& 90.83&\multirow{2}{*}{\textbf{65.63}}\\ 					
 \cline{2-9}
 ~ & 68.91& \textbf{84.85}& 44.09& \textbf{53.6}& \textbf{72.44}& \textbf{43.42}& \textbf{40.73}\\ \hline
 \bottomrule
\end{tabular}}
\label{tb:plusDRM}
\end{table*}

\subsubsection{Bridging Visual Representations for Object Detection in Aerial Images}
\label{sec:AblationBridgingVisualRepresentationsforObjectDetectioninAerialImages}

To evaluate the efficiency of the bridging visual representations module in aerial images, an ablation experiment are designed to compare the BVR module with the baseline in DOTA test dataset. From Table~\ref{tb:plusBVR}, the baseline is the FCOS algorithm the same as Section~\ref{sec:AblationDualRelationshipModule}. +BVR is a combination of the baseline method and the bridging visual representations module.

For 10 categories of plane, baseball diamond (BD), bridge, ground track field (GTF), small vehicle (SV), large vehicle (LV), ship, storage tank (ST), swimming pool (SP) and helicopter (HC), the $AP$ values of the +BVR are higher than values of the baseline by $1.2\%$, $1.98\%$, $1.17\%$, $5.55\%$, $2\%$, $0.31\%$, $0.84\%$, $1.9\%$, $5.01\%$ and $9.8\%$. Thus, +BVR achieves a higher $mAP$ values by $1.67\%$. The value increase of the $mAP$ proves the effectiveness of the BVR method in the field of aerial image detection.

\begin{table*}[]
\caption{Detection accuracy in the ablation study of using BVR module or not with DOTA test dataset. The bold numbers denote the highest values in each class.}
\centering
\resizebox{\textwidth}{!}{
\begin{tabular}{cccccccccccccccccccc}
\toprule
\multirow{2}{*}{\textbf{Method}} & \textbf{Plane}& \textbf{BD}& \textbf{Bridge}& \textbf{GTF}& \textbf{SV}& \textbf{LV}& \textbf{Ship}& \textbf{TC} & \multirow{2}{*}{\textbf{mAP(\%)}}\\
\cline{2-9}
~ & \textbf{BC} & \textbf{ST}& \textbf{SBF}& \textbf{RA}& \textbf{Harbor}& \textbf{SP}& \textbf{HC}\\
\midrule
\multirow{2}*{baseline} & 88.12& 70.77& 44.04& 47.46& 76.36& 65.34& 77.96& 90.83&\multirow{2}{*}{64.25}\\ 					
\cline{2-9}
~ & \textbf{74.31}& 78.37& \textbf{48.3}& \textbf{52.62}& \textbf{72.25}& 42.77& 34.27\\ \hline
\multirow{2}*{+BVR} & \textbf{89.32}& \textbf{72.75}& \textbf{45.21}& \textbf{53.01}& \textbf{78.36}& \textbf{65.65}& \textbf{78.74}& 90.83&\multirow{2}{*}{\textbf{65.92}}\\ 					
\cline{2-9}
~ & 72.88& \textbf{80.27}& 45.42& 52.36& 72.12& \textbf{47.78}& \textbf{44.07}\\ \hline
\bottomrule
\end{tabular}}
\label{tb:plusBVR}
\end{table*}

\subsection{Comparison with the State-of-the-Art}
\label{sec:ComparisonwiththeState-of-the-Art}

To examine and evaluate the performance of the proposed framework RelationRS, the proposed framework is compared with the state-of-the-art algorithms with the DOTA test dataset. Table~\ref{tb:RelationRS} shows the AP and the mAP values of different algorithms.

\begin{table*}[]
\caption{Comparisons with the state-of-the-art single-stage-based detectors in the DOTA1.0 test dataset with horizontal bounding boxes. The baseline is the FCOS algorithm. +DRM means the combination of the FCOS and the dual relationship module. +BVR is a combination of the baseline method and the bridging visual representations module. And the RelationRS is a network adding the DRM and the BVR module to the baseline detector. The \textcolor[RGB]{255,0,0}{red} numbers and the \textcolor[RGB]{0,0,255}{blue} numbers denote the highest values and the second highest values in each class.}
\centering
\resizebox{\textwidth}{!}{
\begin{tabular}{cccccccccccccccccccc}
\toprule
\multirow{2}{*}{\textbf{Method}} & \textbf{Plane}& \textbf{BD}& \textbf{Bridge}& \textbf{GTF}& \textbf{SV}& \textbf{LV}& \textbf{Ship}& \textbf{TC} & \multirow{2}{*}{\textbf{mAP(\%)}}\\
\cline{2-9}
~ & \textbf{BC} & \textbf{ST}& \textbf{SBF}& \textbf{RA}& \textbf{Harbor}& \textbf{SP}& \textbf{HC}\\
\midrule
\multirow{2}*{YOLOv3-tiny \cite{YOLOv3}} & 61.48& 24.35& 4.3& 15.49& 20.27& 30.22& 26.96& 72&\multirow{2}{*}{25.73}\\					
\cline{2-9}
~ & 26.21& 22.91& 14.05& 7.27& 28.78& 27.07& 4.55\\ \hline
\multirow{2}*{SSD \cite{SSD}} & 57.85& 32.79& 16.14& 18.67& 0.05& 36.93& 24.74& 81.16&\multirow{2}{*}{29.86}\\ 					
\cline{2-9}
~ & 25.1& 47.47& 11.22& 31.53& 14.12& 9.09& 0\\ \hline
\multirow{2}*{YOLOv2 \cite{YOLOv2}} & 76.9& 33.87& 22.73& 34.88& 38.73& 32.02& 52.37& 61.65&\multirow{2}{*}{39.2}\\ 					
\cline{2-9}
~ & 48.54& 33.91& 29.27& 36.83& 36.44& 38.26& 11.61\\ \hline
\multirow{2}*{RetinaNet \cite{Focalloss}} & 78.22& 53.41& 26.38& 42.27& 63.64& 52.63& 73.19& 87.17&\multirow{2}{*}{50.39}\\					
\cline{2-9}
~ & 44.64& 57.99& 18.03& 51& 43.39& \textcolor[RGB]{0,0,255}{56.56}& 7.44\\ \hline
\multirow{2}*{YOLOv3 \cite{YOLOv3}} & 79& \textcolor[RGB]{0,0,255}{77.1}& 33.9& \textcolor[RGB]{0,0,255}{68.1}& 52.8& 52.2& 49.8& 89.9&\multirow{2}{*}{60}\\					
\cline{2-9}
~ & \textcolor[RGB]{255,0,0}{74.8}& 59.2& \textcolor[RGB]{255,0,0}{55.5}& 49& 61.5& 55.9& 41.7\\ \hline
\multirow{2}*{SBL \cite{SBL}} & \textcolor[RGB]{0,0,255}{89.15}& 66.04& \textcolor[RGB]{0,0,255}{46.79}& 52.56& 73.06& 66.13& 78.66& \textcolor[RGB]{255,0,0}{90.85}&\multirow{2}{*}{64.77}\\					
\cline{2-9}
~ & 67.4& 72.22& 39.88& \textcolor[RGB]{0,0,255}{56.89}& 69.58& \textcolor[RGB]{255,0,0}{67.73}& 34.74\\ \hline
\multirow{2}*{$SFFM^{d}$ \cite{SFFM}} & 88.1& \textcolor[RGB]{255,0,0}{82.4}& \textcolor[RGB]{255,0,0}{47.7}& \textcolor[RGB]{255,0,0}{72.9}& 45.9& \textcolor[RGB]{255,0,0}{73.5}& 64.4& 90.4&\multirow{2}{*}{\textcolor[RGB]{0,0,255}{66.3}}\\					
\cline{2-9}
~ & 66.7& 50.1& \textcolor[RGB]{0,0,255}{54}& \textcolor[RGB]{255,0,0}{60.1}& \textcolor[RGB]{255,0,0}{77.8}& 51.7& \textcolor[RGB]{255,0,0}{69.5}\\ \hline
\multirow{2}*{baseline \cite{Fcos}} & 88.12& 70.77& 44.04& 47.46& 76.36& 65.34& 77.96& \textcolor[RGB]{0,0,255}{90.83}&\multirow{2}{*}{64.25}\\ 					
\cline{2-9}
~ & \textcolor[RGB]{0,0,255}{74.31}& 78.37& 48.3& 52.62& 72.25& 42.77& 34.27\\ \hline
\multirow{2}*{+DRM} & 87.95& 71.66& 44.1& 52.48& 78.02& 68.41& \textcolor[RGB]{255,0,0}{82.93}& \textcolor[RGB]{0,0,255}{90.83}&\multirow{2}{*}{65.63}\\ 					
\cline{2-9}
~ & 68.91& \textcolor[RGB]{255,0,0}{84.85}& 44.09& 53.6& 72.44& 43.42& 40.73\\ \hline
\multirow{2}*{+BVR} & \textcolor[RGB]{255,0,0}{89.32}& 72.75& 45.21& 53.01& \textcolor[RGB]{0,0,255}{78.36}& 65.65& 78.74& \textcolor[RGB]{0,0,255}{90.83}&\multirow{2}{*}{65.92}\\ 					
\cline{2-9}
~ & 72.88& 80.27& 45.42& 52.36& 72.12& 47.78& 44.07\\ \hline
\multirow{2}*{RelationRS} & 88.27& 72.96& 45.47& 53.7& \textcolor[RGB]{255,0,0}{79.73}& \textcolor[RGB]{0,0,255}{70.98}& \textcolor[RGB]{0,0,255}{82.38}& \textcolor[RGB]{0,0,255}{90.83}&\multirow{2}{*}{\textcolor[RGB]{255,0,0}{66.81}}\\ 					
\cline{2-9}
~ & 69.86& \textcolor[RGB]{0,0,255}{83.29}& 45.26& 54.61& \textcolor[RGB]{0,0,255}{72.79}& 47.85& \textcolor[RGB]{0,0,255}{44.18}\\ \hline
\bottomrule
\end{tabular}}
\label{tb:RelationRS}
\end{table*}

As shown in Table~\ref{tb:RelationRS}, the proposed RelationRS achieves the highest mAP value, and it outperfors YOLOv3-tiny \cite{YOLOv3}, SSD \cite{SSD}, YOLOv2 \cite{YOLOv2}, RetinaNet \cite{Focalloss}, YOLOv3 \cite{YOLOv3}, SBL \cite{SBL}, $SFFM^{d}$ \cite{SFFM}, and FCOS (baseline) \cite{Fcos} by $41.08\%$, $36.95\%$, $27.61\%$, $16.42\%$, $6.81\%$, $2.04\%$, and $0.51\%$. For 15 categories of objects, RelationRS obtains the highest mAP value on 1 class (small vehicle) and the second highest mAP value on 6 classes (large vehicle, ship, tennis court, storage tank, harbor, and helicopter).

RelationRS has good detection performance for small objects, such as small vehicle (SV), large vehicle (LV), ship, storage tank (ST), and helicopter (HC), etc. This phenomenon is similar to the one discussed in Section~\ref{sec:AblationDualRelationshipModule}. In addition, the BVR module is also used in the RelationRS method to improve the performance of the detector on images with complex background. Therefore, the overall accuracy of multiple categories is improved to a certain extent. Interestingly, $SFFM^{d}$ obtains the highest mAP value on 7 classes and the second highest mAP value on 1 classes, but its mAP value is slightly lower than the RelationRS method. This indicates that RelationRS has a more stable performance on different categories in DOTA dataset, and proves the generalization of detector, that is, it is easier to promote the RelationRS to specific object categories not included in the current large aerial image datasets. The above experiments fully demonstrate the effectiveness of the proposed method.

\begin{figure*}
	\begin{center}
		\includegraphics[width=0.8\linewidth]{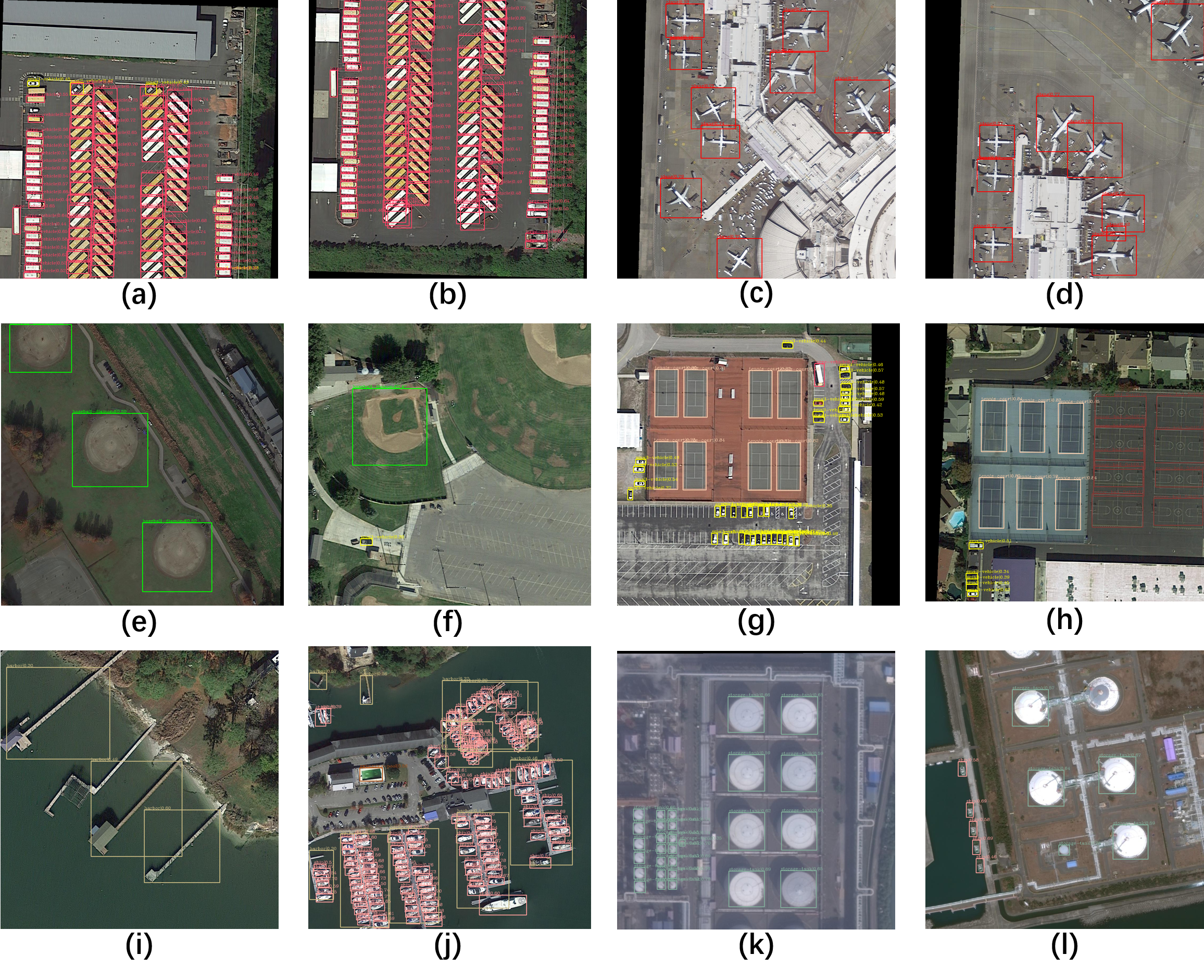}
	\end{center}
	\caption{Object detection results on DOTA1.0 dataset. (a)-(l) are schematic pictures of detection results in different scenarios.}
	\label{results}
\end{figure*}

In conclusion, RelationRS constructs a dual relationship module to guide the fusion of multi-scale features. DRM learns the comparison information of scenes from different patches in one batch and dynamically generates the weight parameters of multi-scale fusion through conditional convolution. Furthermore, the bridging visual representations module for natural image object detection is introduced into RelationRS to improve the performance of the detector on aerial images with complex background information. To the best of our knowledge, this is the first time to demonstrate the performance of BVR Module on aerial image. Some examples of detection results in different scenarios are shown in Figure~\ref{results}.

\section{Conclusions}
\label{sec:Conclusions}

In this paper, a single-stage-based object detector in aerial images is proposed, namely RelationRS. The framework combines a dual relationship module and a bridging visual representations module to solve the problem of multi-scale fusion and improve the object detection accuracy in aerial image with complex background. The dual relationship module can extract the relationship between different scenes according to the input images and dynamically generate the weight parameters required for feature maps fusion. In addition, on the basis of the characteristics of objects in aerial images is not easy to keep out each other, from the viewpoint of introducing key-point detection algorithm, we proves the effectiveness of the bridging visual representations module in the field of aerial image object detection. The experiments under taken with the public DOTA1.0 dataset confirmed the remarkable performance of the proposed method.

On the other hand, single-stage detectors are still not as accurate as two-stage detectors. The current neural networks still cannot be better explained. Therefore, how to better explain the features extracted by neural networks and combine the imaging parameters of aerial images is one of the key points to improve the detection accuracy of aerial image in the future.

\ifCLASSOPTIONcaptionsoff
  \newpage
\fi



%
%
%

\bibliographystyle{IEEEtran}

\begin{thebibliography}{999}
\bibitem[Xia(2018)]{DOTA}
Xia, G.S.; Bai, X.; Zhang, L.P.; Serge, B.; Marcello, P. DOTA: A Large-Scale Dataset for Object Detection in Aerial Images. In Proceedings of the IEEE Conference on Computer Vision and Pattern Recognition, Salt~ Lake City, UT, USA, 18--22 June 2018; pp. 7132--7141.
\bibitem[Li(2020)]{GLSNet}
Li, C.; Luo, B.; Hong, H.; Su, X.; Wang, Y.; Liu, J.; Wang, C.; Zhang, J.; Wei, L. Object Detection Based on Global-Local Saliency Constraint in Aerial Images. {\em Remote Sens}. {\bf 2020}, {\em 12}, 1435. [\href{https://doi.org/10.3390/rs12091435}{CrossRef}]
\bibitem[Li(2020)]{OcSaFPN}
Li, C.; Liu, J.; Hong, H.; Mao, W.; Wang, C.; Hu, C.; Su, X.; Luo, B. Object Detection based on OcSaFPN in Aerial Images with Noise. {\em ArXiv}. {\bf 2020}, arXiv:2012.09859.
\bibitem[Li(2020)]{Lightweight-RS}
Huyan, L.; Bai, Y.; Li, Y.; Jiang, D.; Zhang, Y.; Zhou, Q.; Wei, J.; Liu, J.; Zhang, Y.; Cui, T. A Lightweight Object Detection Framework for Remote Sensing Images. {\em Remote Sens}. {\bf 2021}, {\em 13}, 683. [\href{https://doi.org/10.3390/rs13040683}{CrossRef}]
\bibitem[Liu(2017)]{Optical Satellite Images With Complex Backgrounds}
Liu, Z.; Wang, H.; Weng, L.; Yang, Y. Ship Rotated Bounding Box Space for Ship Extraction From High-Resolution Optical Satellite Images With Complex Backgrounds. {\em IEEE Geosci. Remote Sens. Lett.} {\bf 2017}, {\em 13}, 1074--1078.
\bibitem[Yang(2017)]{Ship Detection From Optical Satellite Images Based on Saliency Segmentation}
Yang, F.; Xu, Q.; Li, B. Ship Detection From Optical Satellite Images Based on Saliency Segmentation and Structure-LBP Feature. {\em IEEE Geosci. Remote Sens. Lett.} {\bf 2017}, {\em 14}, 602--606.
\bibitem[Cheng(2016)]{A Survey}
Cheng, G.; Han, J. A Survey on Object Detection in Optical Remote Sensing Images. {\em ISPRS J. Photogramm. Remote Sens.} {\bf 2016}, {\em 117}, 11--28.
\bibitem[Felzenszwalb(2009)]{DPM}
Felzenszwalb, P.F.; Girshick, R.B.; McAllester, D.; Ramanan, D. Object detection with discriminatively trained part-based models. {\em IEEE Trans. Pattern Anal. Mach. Intell.} {\bf 2009}, {\em 32}, 1627--1645.
\bibitem[Girshick(2014)]{RCNN}
Girshick, R.; Donahue, J.; Darrell, T.; Malik, J. Rich feature hierarchies for accurate object detection and semantic segmentation. In Proceedings of the IEEE Conference on Computer Vision and Pattern Recognition, Columbus, OH, USA, 23--28 June 2014; pp. 580--587.
\bibitem[Girshick(2015)]{FASTRCNN}
Girshick, R. Fast R-CNN. In Proceedings of the IEEE International Conference on Computer Vision, Santiago, Chile, 7--13 December 2015; pp. 1440--1448.
\bibitem[Ren(2017)]{FASTERRCNN}
Ren, S.; He, K.; Girshick, R.; Sun, J. Faster R-CNN: Towards Real-Time Object Detection with Region Proposal Networks. {\em IEEE Trans. Pattern Anal. Mach. Intell.} {\bf 2017}, {\em 39}, 1137--1149.
\bibitem[Lin(2017)]{FPN}
Lin, T.Y.; Dollár, P.; Girshick, R.; He, K.; Hariharan, B.; Belongie, S. Feature pyramid networks for object detection. In Proceedings of the IEEE Conference on Computer Vision and Pattern Recognition, Honolulu, HI,  USA, 21--26 July 2017; pp. 2117--2125.
\bibitem[Cai(2018)]{Cascade}
Cai, Z.; Vasconcelos, N. Cascade r-cnn: Delving into high quality object detection. In Proceedings of the IEEE Conference on Computer Vision and Pattern Recognition, Salt Lake City, UT, USA, 18--22 June 2018; pp.~ 6154--6162.
\bibitem[He(2017)]{Mask r-cnn}
He, K.; Gkioxari, G.; Dollár, P.; Girshick, R. Mask r-cnn. In Proceedings of the IEEE International Conference on Computer Vision, Venice, Italy, 22--29 October 2017; pp. 2961--2969.
\bibitem[Sermanet(2018)]{Overfeat}
Sermanet, P.; Eigen, D.; Zhang, X.; Mathieu, M.; Fergus, R.; Lecun, Y. Overfeat: Integrated recognition, localization and detection using convolutional networks. In~ Proceedings~ of the International Conference on Learning Representations, Banff, AB, Canada, 14--26 April 2014.
\bibitem[Redmon(2016)]{YOLO}
Redmon, J.; Divvala, S.; Girshick, R.; Farhadi, A. You only look once: Unified, real-time object detection. In Proceedings of the IEEE Conference on Computer Vision and Pattern Recognition, Las Vegas, NV,  USA, 26--30 June 2016; pp. 779--788.
\bibitem[Redmon(2017)]{YOLOv2}
Redmon, J.; Farhadi, A. YOLO9000: better, faster, stronger. In Proceedings of the IEEE Conference on Computer Vision and Pattern Recognition, Honolulu, HI, USA, 21--26 July 2017; pp.7263--7271.
\bibitem[Redmon(2018)]{YOLOv3}
Redmon, J.; Farhadi, A. Yolov3: An incremental improvement. {\em arXiv} {\bf 2018}, arXiv:1804.02767.
\bibitem[Bochkovskiy(2020)]{YOLOv4}
Bochkovskiy, A.; Wang, C.; Liao, H. Yolov4: Optimal speed and accuracy of object detection. {\em arXiv} {\bf 2020}, arXiv:2004.10934.
\bibitem[Ge(2021)]{YOLOX}
Ge, Z.; Liu, S.; Wang, F.; Li, Z.; Sun, J. Yolox: Exceeding yolo series in 2021. {\em arXiv} {\bf 2021}, arXiv:2107.08430.
\bibitem[Liu(2016)]{SSD}
Liu, W.; Anguelov, D.; Erhan, D.; Szegedy, C.; Reed, S.; Fu, C. Y.; Berg, A. C. SSD: Single Shot Multibox Detector. In Proceedings of the European Conference on Computer Vision, Amsterdam, The Netherlands, 8--16 October 2016; pp. 21--37.
\bibitem[Tian(2019)]{Fcos}
Tian, Z.; Shen, C.; Chen, H.; He, T. Fcos: Fully convolutional one-stage object detection. In Proceedings of the IEEE Conference on Computer Vision and Pattern Recognition, CA, USA, 16--20 June 2019; pp. 9627-9636.
\bibitem[Lin(2017)]{Focalloss}
Lin, T. Y.; Goyal, P.; Girshick, R.; He, K.; Dollár, P. Focal loss for dense object detection. In Proceedings of the IEEE International Conference on Computer Vision, Venice, Italy, 22--29 October 2017; pp. 2980--2988.
\bibitem[Law(2018)]{CornerNet}
Law, H.; Deng, J. Cornernet: Detecting objects as paired keypoints. In Proceedings of the European Conference on Computer Vision, Munich, Germany, 8--14 September 2018; pp. 734--750.
\bibitem[Law(2019)]{CornerNet-Lite}
Law, H.; Teng, Y.; Russakovsky, O.; Deng, J. CornerNet-Lite: Efficient Keypoint Based Object Detection. {\em arXiv} {\bf 2019}, arXiv:1904.08900.
\bibitem[Kong(2019)]{FoveaBox}
Kong, T.; Sun, F.; Liu, H.; Jiang, Y.; Shi, J. FoveaBox: Beyond Anchor-based Object Detector. {\em arXiv} {\bf 2019}, arXiv:1904.03797.
\bibitem[Lecun(2015)]{Deep learning}
Lecun, Y.; Bengio, Y.; Hinton, G. Deep learning. {\em Nature} {\bf 2015}, {\em 521}, 436--444.
\bibitem[He(2016)]{ResNet}
He, K.; Zhang, X.; Ren, S.; Sun, J. Deep residual learning for image recognition. In Proceedings of the IEEE Conference on Computer Vision and Pattern Recognition, Las Vegas, NV,  USA, 26--30 June 2016; pp. 770--778.
\bibitem[Dong(2021)]{Ship Detection Visual Attention}
Dong, Y.; Chen, F.; Han, S.; Liu, H. Ship Object Detection of Remote Sensing Image Based on Visual Attention. {\em Remote Sens}. {\bf 2021}, {\em 13}, 3192. [\href{https://doi.org/10.3390/rs13163192}{CrossRef}]
\bibitem[Cheng(2020)]{BVR}
Cheng, Chi.; Wei, F.; Hu, H. Relationnet++: Bridging visual representations for object detection via transformer decoder. {\em arXiv} {\bf 2020}, arXiv:2010.15831.
\bibitem[Zou(2016)]{SVDNet}
Zou, Z.; Shi, Z. Ship detection in spaceborne optical image with SVD networks. {\em IEEE Trans. Geosci. Remote Sens.} {\bf 2017}, {\em 54}, 5832-5845.
\bibitem[Van(2011)]{selective1}
Van de Sande, K.E.; Uijlings, J.R.; Gevers, T.; Smeulders, A.W. Segmentation as selective search for object recognition. In Proceedings of the IEEE International Conference on Computer Vision, Barcelona, Spain, 6--11 November 2011; pp. 7.
\bibitem[Uijlings(2013)]{selective2}
Uijlings, J.R.; Van De Sande, K.E.; Gevers, T.; Smeulders, A. W. Selective search for object recognition. {\em Int. J. Comput. Vis.} {\bf 2013}, {\em 104}, 154--171.
\bibitem[Dong(2019)]{Sig-NMS}
Dong, R.; Xu, D.; Zhao, J.; Jiao, L.; An, J. Sig-NMS-Based Faster R-CNN Combining Transfer Learning for Small Target Detection in VHR Optical Remote Sensing Imagery. {\em IEEE Trans. Geosci. Remote Sens.} {\bf 2019}, {\em 57}, 8534-8545.
\bibitem[Deng(2017)]{Toward fast and accurate}
Deng, Z.; Sun, H.; Zhou, S.; Zhao, J.; Zou, H. Toward fast and accurate vehicle detection in aerial images using coupled region-based convolutional neural networks. {\em IEEE J. Sel. Top. Appl. Earth Obs. Remote Sens.} {\bf 2017}, {\em 10}, 3652-3664.
\bibitem[Xiao(2017)]{Airport detection based on a multiscale}
Xiao, Z.; Gong, Y.; Long, Y.; Li, D.; Wang, X.; Liu, H. Airport detection based on a multiscale fusion feature for optical remote sensing images. {\em IEEE Geosci. Remote Sens. Lett.} {\bf 2017}, {\em 14}, 1469-1473.
\bibitem[Xu(2017)]{Deformable convnet with aspect ratio}
Xu, Z.; Xu, X.; Wang, L.; Yang, R.; Pu, F. Deformable convnet with aspect ratio constrained nms for object detection in remote sensing imagery. {\em Remote Sens}. {\bf 2017}, {\em 9}, 1312. [\href{https://doi.org/10.3390/rs9121312}{CrossRef}]
\bibitem[Ren(2018)]{Small object based on Faster}
Ren, Y.; Zhu, C.; Xiao, S. Small object detection in optical remote sensing images via modified faster R-CNN. {\em Appl. Sci.} {\bf 2018}, {\em 8}, 813.
\bibitem[Dai(2017)]{Deformable}
Dai, J.; Qi, H.; Xiong, Y.; Li, Y.; Zhang, G.; Hu, H.; Wei, Y. Deformable convolutional networks. In~ Proceedings~ of the IEEE International Conference on Computer Vision, Venice, Italy, 22--29 October 2017; pp. 764--773.
\bibitem[Hamaguchi(2018)]{Effective use of dilated convolutions}
Hamaguchi, R.; Fujita, A.; Nemoto, K.; Imaizumi, T.; Hikosaka, S. Effective use of dilated convolutions for segmenting small object instances in remote sensing imagery. In Proceedings of the 2018 IEEE winter conference on applications of computer vision, Nevada, USA, 12--15 March 2018; pp. 1442-1450.
\bibitem[Yu(2015)]{dilated convolution}
Yu, F.; Vladlen, K. Multi-scale context aggregation by dilated convolutions. {\em arXiv} {\bf 2015}, arXiv:1511.07122.
\bibitem[Wu(2020)]{CDD-Net}
Wu, Y.; Zhang, K.; Wang, J.; Wang, Y.; Wang, Q.; Li, Q. CDD-Net: A Context-Driven Detection Network for Multiclass Object Detection. {\em IEEE Geosci. Remote Sens. Lett.} {\bf 2020}.
\bibitem[Qiu(2019)]{A2RMNet}
Qiu, H.; Li, H.; Wu, Q.; Meng, F.; Ngan, K.N.; Shi, H. A2RMNet: Adaptively Aspect Ratio Multi-Scale Network for Object Detection in Remote Sensing Images. {\em Remote Sens}. {\bf 2019}, {\em 11}, 1594. [\href{https://doi.org/10.3390/rs11131594}{CrossRef}]
\bibitem[Jiang(2017)]{R2CNN}
Jiang, Y.; Zhu, X.; Wang, X.; Yang, S.; Li, W.; Wang, H.; Fu, P.; Luo, Z. R2CNN: Rotational Region CNN for Orientation Robust Scene Text Detection. {\em arXiv} {\bf 2017}, arXiv:1706.09579.
\bibitem[Ma(2018)]{RRPN}
Ma, J.Q.; Shao, W.Y.; Ye, H.; Wang, L.; Wang, H.; Zheng, Y.B.; Xue, X.Y. Arbitrary-oriented scene text detection via rotation proposals. {\em IEEE Trans. Multimedia} {\bf 2018}, {\em 20}, 3111--3122.
\bibitem[Li(2017)]{Rotation-insensitive}
Li, K.; Cheng, G.; Bu, S.; You, X. Rotation-insensitive and context-augmented object detection in remote sensing images. {\em IEEE Trans. Geosci. Remote Sens.} {\bf 2017}, {\em 56}, 2337-2348.
\bibitem[Ding(2019)]{RoITransformer}
Ding, J.; Xue, N.; Long, Y.; Xia, G.X.; Lu, Q.K. Learning RoI Transformer for Oriented Object Detection in Aerial Images. In Proceedings of the IEEE Conference on Computer Vision and Pattern Recognition, CA, USA, 16--20 June 2019; pp. 2844--2853.
\bibitem[Yang(2019)]{Scrdet}
Yang, X.; Yang, J.; Yan, J.; Zhang, Y.; Zhang, T.; Guo, Z.; Sun, X.; Fu K. Scrdet: Towards more robust detection for small, cluttered and rotated objects. In Proceedings of the IEEE Conference on Computer Vision and Pattern Recognition, CA, USA, 16--20 June 2019; pp. 8232-8241.
\bibitem[Zhang(2019)]{CAD-Net}
Zhang, G.; Lu, S.; Zhang, W. CAD-Net: A context-aware detection network for objects in remote sensing imagery. {\em IEEE Trans. Geosci. Remote Sens.} {\bf 2019}, {\em 57}, 10015-10024.
\bibitem[Li(2020)]{RADet}
Li, Y.; Huang, Q.; Pei, X.; Jiao, L.; Shang, R. RADet: Refine Feature Pyramid Network and Multi-Layer Attention Network for Arbitrary-Oriented Object Detection of Remote Sensing Images. {\em Remote Sens}. {\bf 2020}, {\em 12}, 389. [\href{https://doi.org/10.3390/rs12030389}{CrossRef}]
\bibitem[Wang(2019)]{Mask OBB}
Wang, J.; Ding, J.; Guo, H.; Cheng, W.; Pan, T.; Yang, W. Mask OBB: A Semantic Attention-Based Mask Oriented Bounding Box Representation for Multi-Category Object Detection in Aerial Images. {\em Remote Sens}. {\bf 2019}, {\em 11}, 2930. [\href{https://doi.org/10.3390/rs11242930}{CrossRef}]
\bibitem[Szegedy(2017)]{Inception-v4}
Szegedy, C.; Ioffe, S.; Vanhoucke, V.; Alemi, A. A. Inception-v4, inception-resnet and the impact of residual connections on learning. In Proceedings of the AAAI Conference on Artificial Intelligence, San Francisco, CA, USA, 4--9 February 2017; pp. 4278--4284.
\bibitem[Li(2019)]{Learning object-wise semantic}
Li, C.; Xu, C.; Cui, Z.; Wang, D.; Jie, Z.; Zhang, T.; Yang, J. Learning object-wise semantic representation for detection in remote sensing imagery. In Proceedings of the IEEE Conference on Computer Vision and Pattern Recognition, CA, USA, 16--20 June 2019; pp. 20-27.
\bibitem[Xu(2020)]{Gliding vertex}
Xu, Y.; Fu, M.; Wang, Q.; Wang, Y.; Chen, K.; Xia, G.; Bai, X. Gliding vertex on the horizontal bounding box for multi-oriented object detection. {\em IEEE Trans. Pattern Anal. Mach. Intell.} {\bf 2020}, {\em 43}, 1452-1459.
\bibitem[Zhu(2020)]{Adaptive period embedding}
Zhu, Y.; Du, J.; Wu, X. Adaptive period embedding for representing oriented objects in aerial images. {\em IEEE Trans. Geosci. Remote Sens.} {\bf 2020}, {\em 58}, 7247-7257.
\bibitem[Fu(2020)]{Rotation-aware and multi-scale}
Fu, K.; Chang, Z.; Zhang, Y.; Xu, G.; Zhang, K.; Sun, X. Rotation-aware and multi-scale convolutional neural network for object detection in remote sensing images. {\em ISPRS J. Photogramm. Remote Sens.} {\bf 2020}, {\em 161}, 294-308.
\bibitem[Han(2021)]{Redet}
Han, J.; Ding, J.; Xue, N.; Xia, G. Redet: A rotation-equivariant detector for aerial object detection. In Proceedings of the IEEE Conference on Computer Vision and Pattern Recognition, Virtual, 19--25 June 2021; pp. 2786-2795.
\bibitem[Van(2018)]{YOLT}
Van Etten, A. You only look twice: Rapid multi-scale object detection in satellite imagery. {\em arXiv} {\bf 2018}, arXiv:1805.09512.
\bibitem[Wang(2019)]{FMSSD}
Wang, P.; Sun, X.; Diao, W.; Fu, K. FMSSD: Feature-merged single-shot detection for multiscale objects in large-scale remote sensing imagery. {\em IEEE Trans. Geosci. Remote Sens.} {\bf 2019}, {\em 58}, 3377-3390.
\bibitem[Zou(2017)]{Random access memories}
Zou Z.; Shi Z. Random access memories: A new paradigm for target detection in high resolution aerial remote sensing images. {\em IEEE Trans. Image Process.} {\bf 2017}, {\em 27}, 1100-1111.
\bibitem[Yang(2019)]{R3det}
Yang, X.; Liu, Q.; Yan, J.; Li, A.; Zhang, Z.; Yu, G. R3det: Refined single-stage detector with feature refinement for rotating object. {\em arXiv} {\bf 2019}, arXiv:1908.05612.
\bibitem[Liu(2018)]{PANet}
Liu, S.; Qi, L.; Qin, H.; Shi, J.; Jia, J. Path aggregation network for instance segmentation. In Proceedings of the IEEE Conference on Computer Vision and Pattern Recognition, Salt Lake City, UT, USA, 18--22 June 2018; pp.~ 8759-8768.
\bibitem[Liu(2019)]{ASFF}
Liu, S.; Huang, D.; Wang, Y. Learning spatial fusion for single-shot object detection. {\em arXiv} {\bf 2019}, arXiv:1911.09516.
\bibitem[Liu(2018)]{STDL}
Liu, S.; Qi, L.; Qin, H.; Shi, J.; Jia, J. Path aggregation network for instance segmentation. In Proceedings of the IEEE Conference on Computer Vision and Pattern Recognition, Salt Lake City, UT, USA, 18--22 June 2018; pp. 8759-8768.
\bibitem[Kong(2018)]{Deep feature pyramid reconfiguration}
Kong, T.; Sun, F.; Tan, C.; Liu, H.; Huang, W. Deep feature pyramid reconfiguration for object detection. In Proceedings of the European Conference on Computer Vision, Munich, Germany, 8--14 September 2018; pp. 169-185.
\bibitem[Guo(2020)]{Augfpn}
Guo, C.; Fan, B.; Zhang, Q.; Xiang, S.; Pan, C. Augfpn: Improving multi-scale feature learning for object detection. In Proceedings of the IEEE Conference on Computer Vision and Pattern Recognition, Virtual, 14--29 June 2020; pp. 12595-12604.
\bibitem[Wang(2021)]{U2-ONet}
Wang, C.; Li, C.; Liu, J.; Luo, B.; Su, X.; Wang, Y.; Gao, Y. U2-ONet: A Two-Level Nested Octave U-Structure Network with a Multi-Scale Attention Mechanism for Moving Object Segmentation. {\em Remote Sens}. {\bf 2021}, {\em 13}, 60. [\href{https://doi.org/10.3390/rs13010060}{CrossRef}]
\bibitem[Ghiasi(2019)]{Nas-fpn}
Ghiasi, G.; Lin, T.; Le, Q. Nas-fpn: Learning scalable feature pyramid architecture for object detection. In Proceedings of the IEEE Conference on Computer Vision and Pattern Recognition, CA, USA, 16--20 June 2019; pp. 7036-7045.
\bibitem[Chen(2020)]{Mnasfpn}
Chen, B.; Ghiasi, G.; Liu, H.; Lin, T.; Kalenichenko, D.; Adam, H.; Le, Q. Mnasfpn: Learning latency-aware pyramid architecture for object detection on mobile devices. In Proceedings of the IEEE Conference on Computer Vision and Pattern Recognition, Virtual, 14--29 June 2020; pp. 13607-13616.
\bibitem[Tan(2020)]{BiFPN}
Tan, M.; Pang, R.; Le, Q. Efficientdet: Scalable and efficient object detection. In Proceedings of the IEEE Conference on Computer Vision and Pattern Recognition, Virtual, 14--29 June 2020; pp. 10781-10790.
\bibitem[Jia(2016)]{Dynamic filter}
Jia, X.; De Brabandere, B.; Tuytelaars, T.; Gool, L. Dynamic filter networks. {\em Adv. Neural Inf Process Syst.} {\bf 2016}, {\em 29}, 667-675.
\bibitem[Ha(2016)]{Hypernetworks}
Ha, D.; Dai, A.; Le, Q. Hypernetworks. {\em arXiv} {\bf 2016}, arXiv:1609.09106.
\bibitem[Shen(2018)]{Neural style transfer}
Shen, F.; Yan, S.; Zeng, G. Neural style transfer via meta networks. In Proceedings of the IEEE Conference on Computer Vision and Pattern Recognition, Salt Lake City, UT, USA, 18--22 June 2018; pp. 8061-8069.
\bibitem[Jo(2018)]{Deep video super-resolution}
Jo, Y.; Oh, S. W.; Kang, J.; Kim, S. Deep video super-resolution network using dynamic upsampling filters without explicit motion compensation. In Proceedings of the IEEE Conference on Computer Vision and Pattern Recognition, Salt Lake City, UT, USA, 18--22 June 2018; pp. 3224-3232.
\bibitem[Hu(2019)]{Meta-SR}
Hu, X.; Mu, H.; Zhang, X.; Wang, Z.; Tan, T.; Sun, J. Meta-SR: A magnification-arbitrary network for super-resolution. In Proceedings of the IEEE Conference on Computer Vision and Pattern Recognition, CA, USA, 16--20 June 2019; pp. 1575-1584.
\bibitem[Yang(2019)]{Condconv}
Yang, B.; Bender, G.; Le, Q.; Ngiam, J. Condconv: Conditionally parameterized convolutions for efficient inference. {\em arXiv} {\bf 2019}, arXiv:1904.04971.
\bibitem[Wu(2018)]{Dynamic filtering with large sampling}
Wu, J.; Li, D.; Yang, Y.; Bajaj, C.; Ji, X. Dynamic filtering with large sampling field for convnets. In Proceedings of the European Conference on Computer Vision, Munich, Germany, 8--14 September 2018; pp. 185-200.
\bibitem[Harley(2017)]{Segmentation-aware convolutional networks}
Harley, A.; Derpanis, K.; Kokkinos, I. Segmentation-aware convolutional networks using local attention masks. In~ Proceedings~ of the IEEE International Conference on Computer Vision, Venice, Italy, 22--29 October 2017; pp. 5038-5047.
\bibitem[Tian(2018)]{CondInst}
Tian, Z.; Shen, C.; Chen, H. Conditional convolutions for instance segmentation. In Proceedings of the European Conference on Computer Vision, Glasgow, UK, 23--28 August 2018; pp. 282-298.
\bibitem[Xue(2016)]{Visual dynamics}
Xue, T.; Wu, J.; Bouman, K. L.; Freeman, W. Visual dynamics: Probabilistic future frame synthesis via cross convolutional networks. {\em arXiv} {\bf 2016}, arXiv:1607.02586.
\bibitem[Sagong(2019)]{cGANs}
Sagong, M.; Shin, Y.; Yeo, Y.; Park, S.; Ko, S. cGANs with Conditional Convolution Layer. {\em arXiv} {\bf 2019}, arXiv:1906.00709.
\bibitem[Liu(2019)]{Learning to predict layout-to-image}
Liu, X.; Yin, G.; Shao, J.; Wang, X.; Li, H. Learning to predict layout-to-image conditional convolutions for semantic image synthesis. {\em arXiv} {\bf 2019}, arXiv:1910.06809.
\bibitem[Liu(2021)]{CondLaneNet}
Liu, L.; Chen, X.; Zhu, S.; Tan, P. CondLaneNet: a Top-to-down Lane Detection Framework Based on Conditional Convolution. {\em arXiv} {\bf 2021}, arXiv:2105.05003.
\bibitem[Yang(2021)]{ConDinet++}
Yang, K.; Yi, J.; Chen, A.; Liu, J.; Chen, W. ConDinet++: Full-Scale Fusion Network Based on Conditional Dilated Convolution to Extract Roads From Remote Sensing Images. {\em IEEE Geosci. Remote Sens. Lett.} {\bf 2021}.
\bibitem[Krizhevsky(2012)]{ImageNet Classification}
Krizhevsky, A.; Sutskever, I.; Hinton, G. ImageNet Classification with Deep Convolutional Neural Networks. In Proceedings of the International Conferenceon Neural Information Processing Systems, Lake Tahoe, ND, USA, 3--8 December 2012; pp. 1097--1105.
\bibitem[Chen(2019)]{MMDetection}
Chen, K.; Wang, J.; Pang, J.; Cao, Y.; Xiong, Y.; Li, X.; Sun, S.; Feng, W.; Liu, Z.; Xu, J.; Zhang, Z.; Cheng, D.; Zhu, C.; Cheng, T.; Zhao, Q.; Li, B.; Lu, X.; Zhu, R.; Wu, Y.; Dai, J.; Wang, J.; Shi, J.; Ouyang, W.; Loy, C.; Lin, D. MMDetection: Open mmlab detection toolbox and benchmark. {\em arXiv} {\bf 2019}, arXiv:1906.07155.
\bibitem[Sutskever(2012)]{AlexNet}
Sutskever, I.; Hinton, G.E.; Krizhevsky, A. Imagenet classification with deep convolutional neural networks. {\em Adv. Neural Inf. Process. Syst.} {\bf 2012}, \emph{60}, 1097--1105.
\bibitem[Vaswani(2017)]{Transformer}
Vaswani, A.; Shazeer, N.; Parmar, N.; Uszkoreit, J.; Jones, L.; Gomez, A.; Kaiser. L.; Polosukhin, I. Attention is all you need. In Proceedings of the International Conference on Neural Information Processing Systems, Long Beach, CA, USA, 4--9 December 2017; pp. 5998-6008.
\bibitem[Sun(2018)]{SBL}
Sun, P.; Chen, G.; Luke, G.; Shang, Y. Salience biased loss for object detection in aerial images. {\em arXiv} {\bf 2018}, arXiv:1810.08103.
\bibitem[Wang(2019)]{SFFM}
Wang, P.; Sun, X.; Diao, W.; Fu, K. Mergenet: Feature-merged network for multi-scale object detection in remote sensing images. In Proceedings of the IEEE International Geoscience and Remote Sensing Symposium, Yokohama, Japan, 28--2 July--August 2019; pp. 238-241.
\end{thebibliography}


%

%
%
%




\end{document}